\documentclass[letterpaper]{article} 
\usepackage{aaai2026}  
\usepackage{times}  
\usepackage{helvet}  
\usepackage{courier}  
\usepackage[hyphens]{url}  
\usepackage{graphicx} 
\usepackage{natbib}  
\usepackage{caption} 
\usepackage{algorithm}
\usepackage{algorithmic}

\usepackage{newfloat}
\usepackage{listings}
\DeclareCaptionStyle{ruled}{labelfont=normalfont,labelsep=colon,strut=off} 
\floatstyle{ruled}
\newfloat{listing}{tb}{lst}{}
\floatname{listing}{Listing}
\usepackage{bm}
\usepackage{multirow}
\usepackage{colortbl}
\usepackage{makecell}
\usepackage{subcaption}
\usepackage{amsmath}
\usepackage{color}
\usepackage{amsfonts}
\usepackage{arydshln}
\usepackage{pifont}
\usepackage{dashbox}
\usepackage{xspace}
\usepackage{siunitx}  
\usepackage{booktabs} 
\usepackage{array}    
\usepackage{enumitem}

\newcommand{\ours}{{RF-CLIP}}
\newcommand{\argmax}{\mathop{\arg\max}}
\newcommand{\argmin}{\mathop{\arg\min}}
\newcommand{\softmax}{\mathrm{softmax}}
\newcommand{\attn}{\operatorname{Attn}}

\usepackage[dvipsnames,table,xcdraw]{xcolor}

\title{Target Refocusing via Attention Redistribution for Open-Vocabulary Semantic Segmentation: An Explainability Perspective}
\author {
    Jiahao Li\textsuperscript{\rm 1},
    Yang Lu\textsuperscript{\rm 1},
    Yachao Zhang\textsuperscript{\rm 1},
    Yong Xie\textsuperscript{\rm 3}\(^\ast\),\\
    Fangyong Wang\textsuperscript{\rm 4},
    Yuan Xie\textsuperscript{\rm 2},
    Yanyun Qu\textsuperscript{\rm 1}\thanks{Corresponding author. \\Code: https://github.com/liblacklucy/RF-CLIP}
}
\affiliations {
    \textsuperscript{\rm 1}Key Laboratory of Multimedia Trusted Perception and Eficient Computing.Ministry of Education of China, School of Informatics, Xiamen University\\
    \textsuperscript{\rm 2}School of Computer Science and Technology, East China Normal University\\
    \textsuperscript{\rm 3}Department of Computer, Nanjing University of Posts and Telecommunications\\
    \textsuperscript{\rm 4}Hanjiang National Laboratory\\
}

\usepackage{bibentry}

\begin{document}
\begin{sloppypar}

\maketitle

\begin{abstract}
Open-vocabulary semantic segmentation (OVSS) employs pixel-level vision-language alignment to associate category-related prompts with corresponding pixels. A key challenge is enhancing the multimodal dense prediction capability, specifically this pixel-level multimodal alignment. Although existing methods achieve promising results by leveraging CLIP’s vision-language alignment, they rarely investigate the performance boundaries of CLIP for dense prediction from an interpretability mechanisms perspective. In this work, we systematically investigate CLIP's internal mechanisms and identify a critical phenomenon: analogous to human distraction, CLIP diverts significant attention resources from target regions to irrelevant tokens. Our analysis reveals that these tokens arise from dimension-specific over-activation; filtering them enhances CLIP's dense prediction performance. Consequently, we propose \underline{R}e\underline{F}ocusing CLIP (\ours), a training-free approach that emulates human distraction-refocusing behavior to redirect attention from distraction tokens back to target regions, thereby refining CLIP's multimodal alignment granularity. Our method achieves SOTA performance on eight benchmarks while maintaining high inference efficiency.
\end{abstract}


\section{Introduction}
Open-vocabulary semantic segmentation assigns categorical labels from an unrestricted vocabulary to each image pixel via prompt-guided multimodal semantic alignment. The core challenge lies in achieving pixel-level dense prediction, where fine-grained multimodal alignment constitutes a critical bottleneck. Existing approaches predominantly leverage CLIP's~\cite{radford2021learning} vision-language alignment capability to address this challenge, falling into three paradigms: 1) \textbf{Joint Fine-tuning}~\cite{cho2024cat,jiao2023learning,li2024relationship}: simultaneously fine-tuning CLIP alongside segmentation-specific components to enhance dense prediction capabilities; 2) \textbf{Pre Fine-tuning}~\cite{wu2023clipself,wu2024clim,xu2022groupvit}: refining CLIP's alignment granularity through fine-grained vision-language contrastive learning; 3) \textbf{Training-Free Adaptation}\cite{lan2024clearclip,wang2024sclip,lan2024proxyclip}: modulating CLIP's final residual attention layer or integrating vision foundation models (VFMs)\cite{zhang2022dino,oquab2023dinov2} to boost alignment granularity. Collectively, these paradigms deploy complementary strategies—augmenting CLIP with task-specific modules, re-pretraining weights, or adapting final layers while aggregating VFMs—all targeting enhanced multimodal alignment granularity. Despite promising results, existing methods rarely investigate the performance boundaries of CLIP for dense prediction from an interpretability perspective or explore the origins of its inherent inter-layer spatial misalignment, thereby limiting further improvements in OVSS performance.

\begin{figure}[t]
  \includegraphics[width=\linewidth]{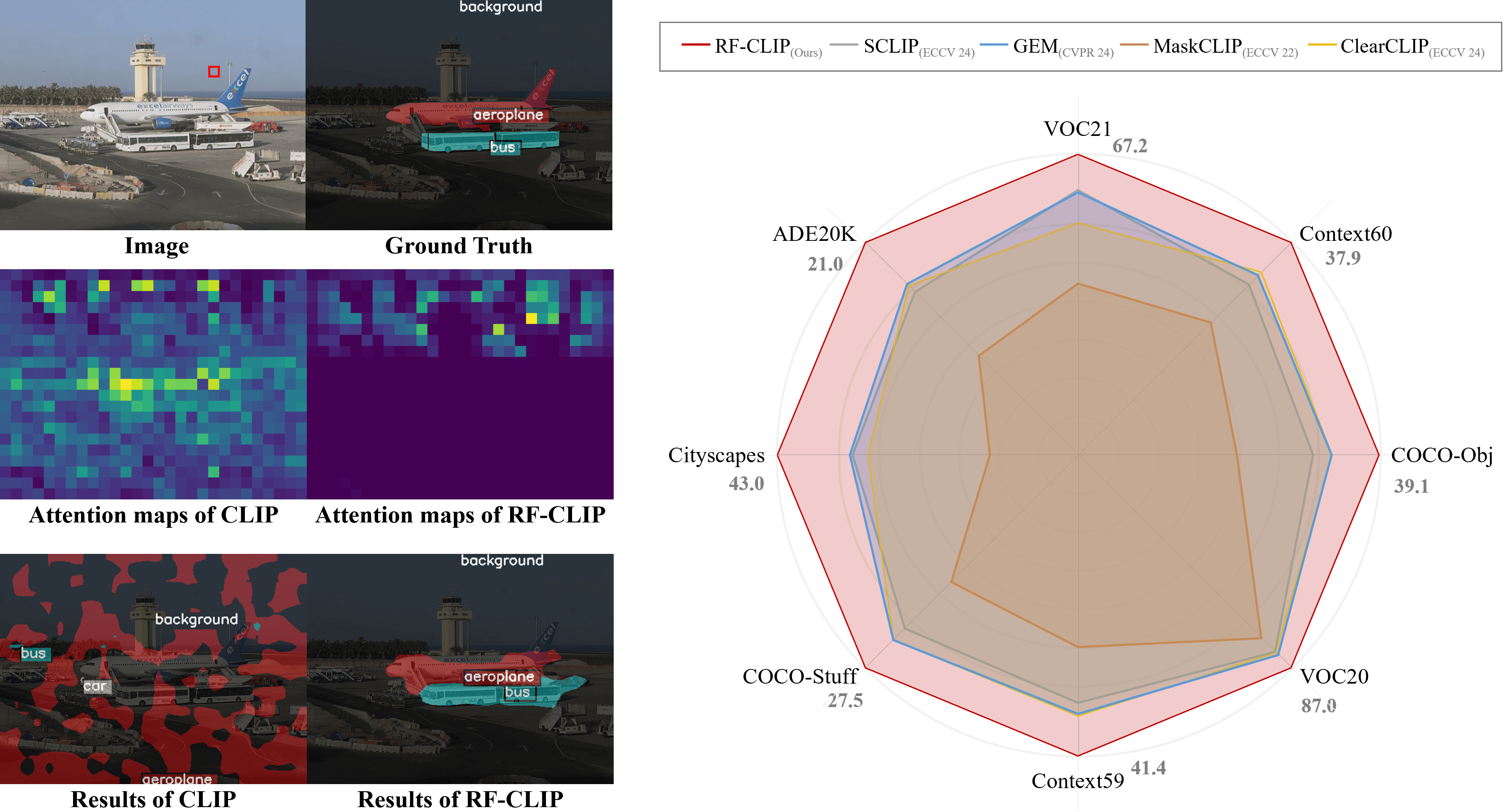}
  \caption{RF-CLIP achieves precise attention focus, which facilitates accurate segmentation of target regions. Compared to state-of-the-art methods, RF-CLIP demonstrates superior performance, achieving the highest accuracy.}
  \label{fig:intro}
\end{figure}

In this work, we focus on the training-free paradigm, exclusively modulating CLIP to enhance pixel-level dense prediction. We first systematically investigate CLIP's internal mechanisms and uncover a key phenomenon: during visual encoding, CLIP produces certain tokens irrelevant to the input query that consume substantial attention resources, causing distraction from originally focused target regions (termed the ``distraction'' phenomenon causing layer-wise spatial misalignment). Further analysis reveals these tokens stem from over-activation in specific dimensions; filtering them significantly improves CLIP's OVSS performance. Their presence seems to indicate that CLIP struggles to focus on the current visual targets, impairing its dense prediction ability. This insight motivates us to explore whether mimicking the human distraction-refocusing behavior — by reallocating attention resources from distraction tokens to defocused target regions — can enhance CLIP's multimodal alignment granularity.

To this end, we propose \underline{R}e\underline{F}ocusing CLIP (\ours), a straightforward training-free modulation method that operates directly on CLIP's inter-layer attention mechanisms. The core idea lies in: 1) identifying query-irrelevant tokens consuming substantial attention resources and locating defocused target regions within CLIP's intermediate visual embeddings; 2) reallocating attention resources from distraction tokens to these defocused target regions. By exclusively modulating CLIP's attention mechanism, \ours \xspace significantly enhances OVSS performance while preserving inference efficiency (Figure~\ref{fig:intro}). Our principal contributions are:
\begin{itemize}[leftmargin=*]
\item We identify the ``distraction'' phenomenon and attribute it to dimension-specific over-activation inherent in CLIP, which undermines the multimodal alignment granularity.
\item We propose \ours, a training-free attention modulator that enhances alignment granularity by reallocating misallocated attention resources to defocused target regions.
\item Our method achieves SOTA performance on eight OVSS benchmark datasets while simultaneously preserving superior computational efficiency during inference.
\end{itemize}


\begin{figure*}[t]
  \includegraphics[width=\textwidth]{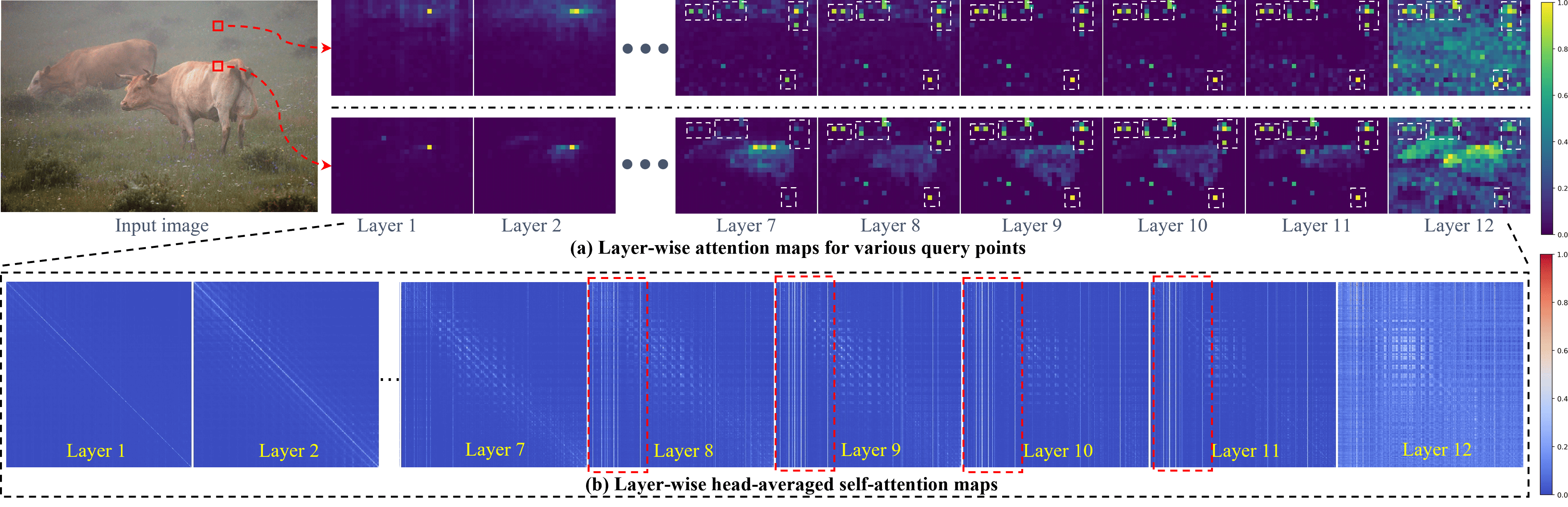}
  \caption{Illustration of ``distraction'' phenomenon. (a) Layer-wise attention maps for two query locations reveal query-to-tokens relevance across the entire image. (b) Layer-wise head-averaged self-attention maps characterize token-to-token relationships. Red solid boxes denote query points; white dashed boxes indicate distraction tokens; and red dashed boxes represent attention weights of distraction tokens.}
  \label{fig:distraction}
\end{figure*}

\section{Preliminaries}

\paragraph{Problem definition.} Given an image ${\bm I}$ and a set of class-specific textual descriptions $\{{\bm T}_i\}^{N_c}_{i=1}$, where ${\bm T_i}$ denotes the textual description of the $i$-th class and $N_c$ represents the total number of classes, CLIP aligns each pixel in ${\bm I}$ with the most semantically relevant ${\bm T_i}$, thereby assigning the corresponding class label $i$ to the pixel. Note that the total number of classes $N_c$ is dynamic during inference.

\paragraph{Dense inference for training-free OVSS.} The ViT-based CLIP model primarily comprises multiple stacked residual attention layers.  Each residual layer contains a self-attention module ${\mathcal{S}}(\cdot)$ and a feedforward module ${\mathcal{F}}(\cdot)$. Given an image ${\bm I}$, a visual residual layer processes it to extract visual embeddings $\bm{f} \in \mathbb{R}^{N\times d}$, where $N$ represents the number of patches and $d$ is the latent space dimension. (In actual, $\bm{f}$ comprises a global visual embeddings $\bm{f}^{'} \in \mathbb{R}^{1\times d}$ and a local dense embeddings $\bm{f}^{''} \in \mathbb{R}^{N\times d}$; for simplicity, we refer to $\bm{f}$ as the local dense embeddings $\bm{f}^{''}$ and ignore the global visual embeddings). The computation process within a visual residual layer is defined as (normalization operations omitted for simplicity):
\begin{align}
  \bm{f}^l &\leftarrow \bm{f}^l + {\mathcal{S}}(\bm{f}^l), \label{eq:1}\\
  \bm{f}^{l+1} &\leftarrow \bm{f}^l + {\mathcal{F}}(\bm{f}^l), \label{eq:2}
\end{align}
where $l\in[1,L]$ denote the current layer index. The computation process of self-attention module ${\mathcal{S}}(\cdot)$ is defined as:
\begin{align}
  \mathcal{S}(\bm{f}^l)&=\attn^l_{qk}\cdot \bm{V}^l,\, \attn^l_{qk}\!=\!\softmax(\frac{\bm{Q}^l{\bm{K}^l}^\top}{\sqrt{d}}),\\
  \bm{Q}^l&=\bm{f}^l\bm{W}^l_{q}, \quad \bm{K}^l=\bm{f}^l\bm{W}^l_{k}, \quad \bm{V}^l=\bm{f}^l\bm{W}^l_{v},
\end{align}
where $\bm{W}_q,\bm{W}_k,\bm{W}_v$ are linear projection matrices, and $\attn_{qk}$ are self-attention maps. Similarly, given a set of textual descriptions $\{{\bm T}_i\}^{N_c}_{i=1}$, CLIP textual encoder extracts its textual embeddings $\bm{f}_t \in \mathbb{R}^{N_c\times d}$. Thus, the final open-vocabulary semantic segmentation map $\bm{M} \in \mathbb{R}^{H\times W}$ (or $\in \mathbb{R}^{N\times 1}$ if flattened) is:
\begin{equation}
  \bm{M} = \argmax_{N_c} \cos{(\bm{f}^L, \bm{f}_t)}.
\end{equation}
In addition, due to the residual connection (including the feedforward module) in the final residual layer often producing ``noisy'' segmentation maps~\cite{lan2024clearclip}, the output of its self-attention module is typically used as the final output visual embeddings $\bm{f}^{L}$. The computation within the last visual residual layer is: $\bm{f}^{L} \leftarrow \mathcal{S}(\bm{f}^{L-1})$.

\paragraph{Training-free adaptation for OVSS.} Omission of the residual connection in the final layer and reliance solely on $\mathcal{S}(\cdot)$ markedly enhance the perceptual granularity of visual embeddings. Consequently, existing research focuses on refining $\mathcal{S}(\cdot)$ to achieve finer-grained spatial alignment. These methods primarily aim to optimize the final self-attention matrices $\attn^L_{q,k} \in \mathbb{R}^{N\times N}$ for more precise modeling of local spatial relationships among image patches. They fall into two main categories:
\begin{itemize}[leftmargin=*]
\item \textbf{VFM-proxy.} These methods leverage the dense representation capabilities of powerful visual foundation models to enhance CLIP. ProxyCLIP~\cite{lan2024proxyclip} utilizes DINO~\cite{zhang2022dino}'s visual representations to compute self-similarity, replacing CLIP's original self-attention matrices. CASS~\cite{kim2025distilling} employs spectral graph distillation to integrate DINO's dense visual features, thereby strengthening CLIP's contextual coherence.
\item \textbf{Self-proxy.} They construct novel self-attention matrices from their own embeddings, replacing the original $\attn^L_{q,k}$. For instance, SCLIP~\cite{wang2024sclip} utilizes the summation of $\attn^L_{qq}$ and $\attn^L_{kk}$ as the final self-attention matrix. ClearCLIP~\cite{lan2024clearclip} and NACLIP~\cite{hajimiri2025pay} respectively substitute $\attn^L_{qk}$ with $\attn^L_{qq}$ and $\attn^L_{kk}$. Later we refer to $\attn_{kk}$-proxy CLIP as CLIP with $\attn^L_{kk}$ replacing $\attn^L_{qk}$.
\end{itemize}
Despite the impressive achievements of the aforementioned paradigms, they exhibit a fundamental limitation: these approaches exclusively modulate the final residual attention layer (to prevent model collapse), while ignoring patch-level spatial misalignments within intermediate residual layers. This oversight causes accumulated error propagation across the network. This raises a critical question: Could direct correction of spatial misalignments in intermediate residual layers enhance CLIP’s dense prediction performance? To this end, in the next section, we systematically analyze the self-attention matrices layer-wise to probe CLIP's underlying mechanisms from an interpretability perspective.

\section{``Distraction'' phenomenon}
In this section, we first delineate the ``distraction'' phenomenon, which reveals pronounced spatial misalignment artifacts within CLIP. Furthermore, our empirical analysis establishes that this phenomenon originates from excessive activation in specific dimensions. Building on this analysis, we precisely identify distraction tokens. Finally, we systematically evaluate diverse suppression strategies on these distraction tokens, thereby inspiring our proposed method.
\paragraph{Key observations.} Figure~\ref{fig:distraction} presents layer-wise attention maps $\attn^l_{qk}[i]$ capturing token relevance to diverse query points $i$, alongside head-averaged self-attention maps revealing token-to-token relationships. Attention maps visualization (Figure~\ref{fig:distraction} (a)) demonstrates that shallow layers (1-2) predominantly attend to query-relevant tokens. In contrast, deeper layers (7-12) exhibit numerous high-attention tokens unrelated to target queries (demarcated by white dashed boxes) - a phenomenon we term "distraction," with these tokens designated as distraction tokens $\mathcal{T}_{dis}$. The emergence of distraction tokens progressively diminishes saliency around query-related regions, defocusing originally concentrated target areas. Crucially, these distraction tokens consistently occupy identical spatial positions across two varying query points, perhaps suggesting their universal correlations with all queries points. Self-attention maps visualization (Figure~\ref{fig:distraction} (b)) corroborate this behavior: distraction tokens manifest distinct vertical stripes (within red dashed boxes) due to uniformly high attention values between all tokens and them, i.e., their attention weights $\sum_j\attn^l_{qk}[j,i]_{,i\in \mathcal{T}_{dis}}$ are large, reflecting the sum of the relevance of them to all tokens. Therefore, these tokens fundamentally induce spatial misalignment, propagating errors through residual layers, and ultimately constraining CLIP's dense prediction ability.

\begin{figure}[t]
  \includegraphics[width=\linewidth]{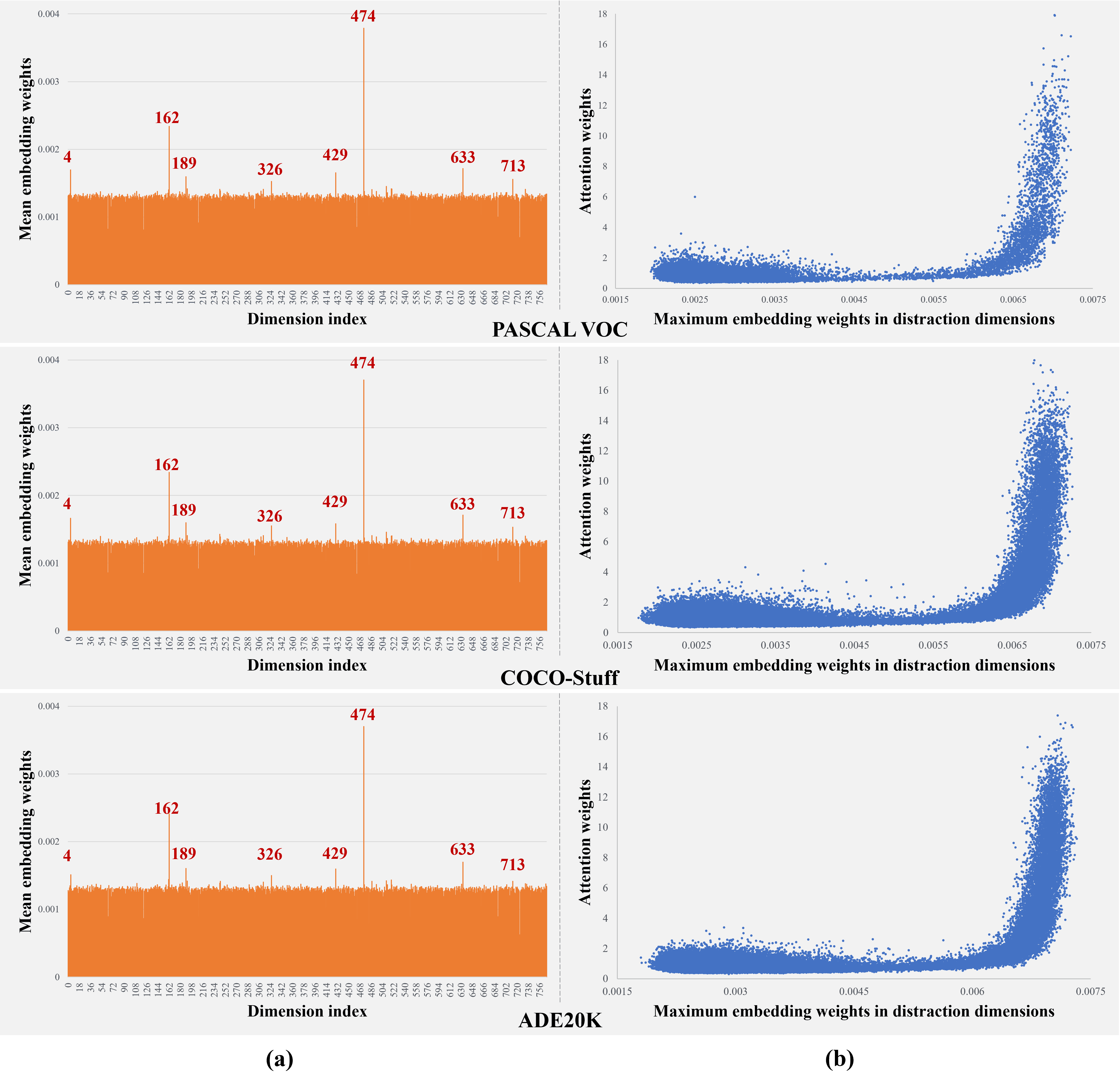}
  \caption{Weights analysis. (a) Histogram of per-dimension mean embedding weights averaged across the entire dataset. (b) Scatter plot of tokens' attention weights versus their maximum embedding weights in distraction dimensions.}
  \label{fig:where}
\end{figure}

\begin{figure}[t]
  \includegraphics[width=\linewidth]{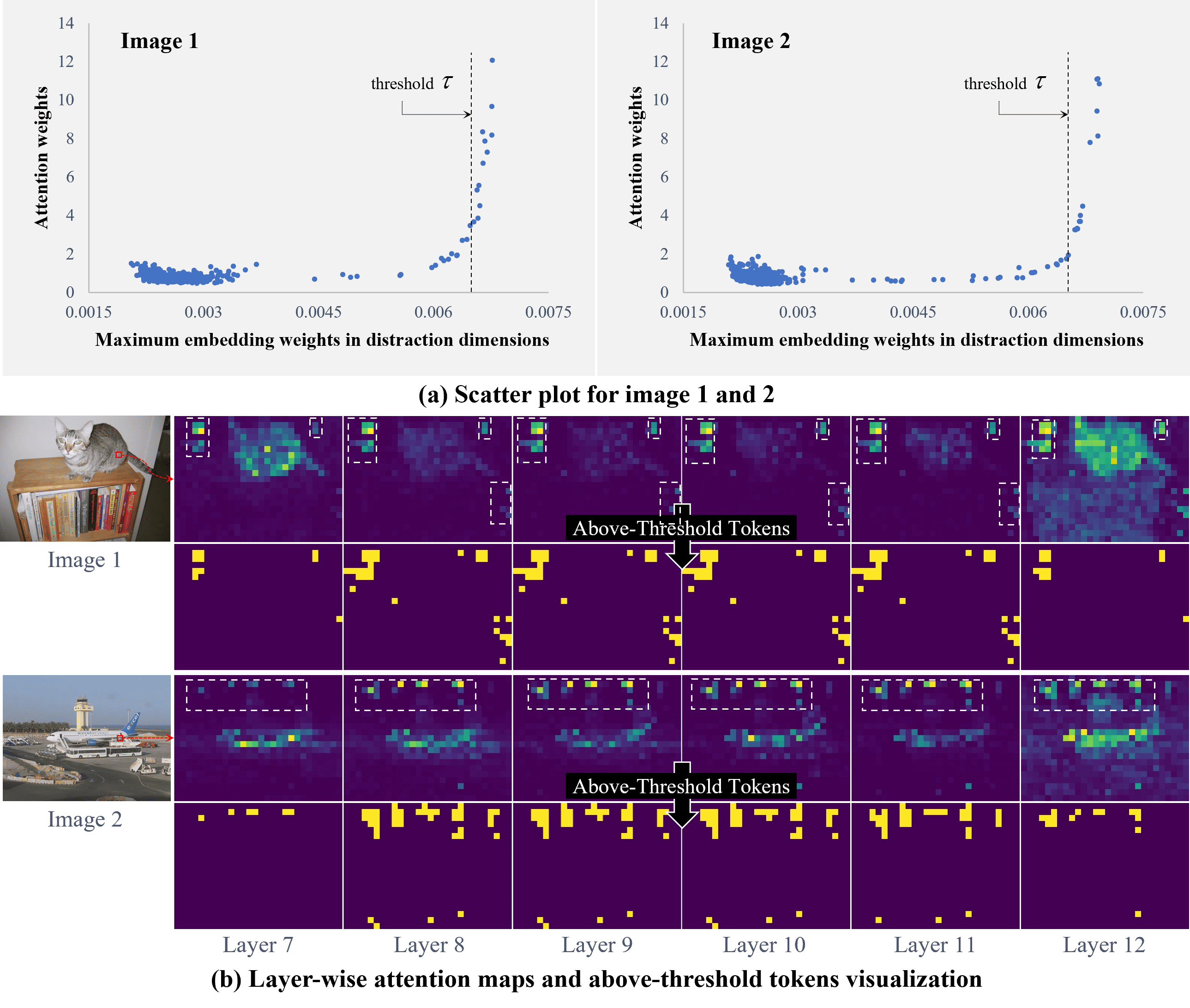}
  \caption{Distraction tokens identification. (a) Scatter plot of attention weights versus maximum embedding weights in $\mathcal{D}_{dis}$. (b) Layer-wise attention maps and above-threshold tokens visualizations.}
  \label{fig:distraction visualization}
\end{figure}

\paragraph{How do distraction tokens surface?} Given distraction tokens' persistent high correlation with all tokens, we hypothesize they exhibit massive embedding weights. We compute visual dense embeddings averaged across residual layers as: $\bm{\bar f}=\frac{1}{L}\sum^L_{l=1}(\frac{\bm{f}^{l}}{\sum^d_{j=1} \bm{f}^{l}[:,j]}) \in \mathbb{R}^{N\times d}$, where $L$ denotes total layers. The embedding weights of $i$-th token in $j$-th dimension can be defined as $\bm{\bar f}[i, j]$, representing its weights distribution across diverse dimensions. Figure~\ref{fig:where} (a) displays the mean embedding weights averaged across the entire dataset, reflecting the weights distribution of the entire dataset in CLIP‘s high-dimensional space. We find that three large-scale OVSS benchmark datasets exhibit consistent weights distribution, with several peaks at identical dimensions (e.g., 4, 162, 474, etc., denoted red), which we designate as distraction dimensions $\mathcal{D}_{dis}$. This indicates CLIP intrinsically generates massive embedding weights in $\mathcal{D}_{dis}$ — an inherent property independent of dataset specifics. Figure~\ref{fig:where} (b) validates our hypothesis, plotting for each token $i$: Attention weights ($\sum_j\attn[j,i]$\footnote{Here, $\attn=\frac{1}{L\cdot H}\sum^{l,h=L,H}_{l,h=1}\attn^{l,h}_{qk}$ denotes self-attention matrices averaged across all $L$ layers and $H$ heads.}) and Maximum embedding weights $\phi_i$ in $\mathcal{D}_{dis}$ ($\phi_i=\max_{j\in \mathcal{D}_{dis}} \bm{\bar f}[i, j]$). We observe that tokens with massive $\phi_i$ consistently exhibit high attention weights, and that growth in $\phi_i$ causes exponential growth in attention weights. Therefore, we conclude that \textbf{tokens with massive maximum embedding weights in the distraction dimensions inevitably evolve into distraction tokens under self-attention calculation mechanisms}.
\paragraph{Where are distraction tokens?} Building upon the above analysis, we define $\tau=5/d$ as the identification threshold, where token $i$ satisfying $\phi_i > \tau$ are identified as distraction tokens. As shown in Figure~\ref{fig:distraction visualization} (a), the scatter plot maps of two images show that above-threshold tokens exhibit high attention weights. Figure~\ref{fig:distraction visualization} (b) visually confirms these tokens' spatial distribution, revealing near-perfect alignment with observed distraction patterns. Therefore, distraction token identification reduces to a simple threshold operation on maximum embedding weights within $\mathcal{D}_{dis}$.

\paragraph{Suppression strategies on distraction tokens.} To explore whether suppressing distraction tokens enhances CLIP's dense prediction capability, we establish an $\attn_{kk}$-proxy CLIP baseline and evaluate four suppression strategies: $-\infty$ masking\footnote{$-\infty$ forces softmax-attention value to zero} - nullifying layer-wise distraction tokens embeddings, low-pass filtering - clamping these embeddings to $\tau$, mean filtering - replacing with mean of $3\times3$ local neighborhoods, and median filtering - substituting with median of $3\times3$ neighborhoods. As shown in Table~\ref{tab:suppressing}, both $-\infty$ masking and low-passing filtering degrade performance, whereas mean and median filtering yield measurable improvements. This indicates that distraction tokens should maintain spatial consistency with adjacent regions: Direct elimination disrupts the topological structure of CLIP's high-dimensional space, inducing model collapse. Consequently, we are motivated to investigate whether reallocating attention resources from distraction tokens to defocused target regions can enhance CLIP's dense prediction capability.

\begingroup
\renewcommand{\arraystretch}{1.0}
\begin{table}[t]
\centering
\resizebox{\linewidth}{!}{
    \begin{tabular}{l!{\vrule height 10pt}cccccc}
    \toprule
    Method & VOC21 & COCO-Stuff & Cityscapes & ADE20k & Avg.\\
    \midrule
    \midrule
    \rowcolor{gray!5}
    baseline & 58.1 & 23.0 & 31.1 & 16.3 & 32.1 \\
    \hdashline
    $-\infty$ masking & 3.5 {\scriptsize \textcolor{red}{(-54.6)}} & 0.1 {\scriptsize \textcolor{red}{(-22.9)}} & 2.0 {\scriptsize \textcolor{red}{(-29.1)}} & 0.1 {\scriptsize \textcolor{red}{(-16.2)}} & 1.4 \\
    low-pass filtering & 7.9 {\scriptsize \textcolor{red}{(-50.2)}} & 1.1 {\scriptsize \textcolor{red}{(-21.9)}} & 6.2 {\scriptsize \textcolor{red}{(-24.9)}} & 1.4 {\scriptsize \textcolor{red}{(-14.9)}} & 4.2 \\
    mean filtering & 59.3 {\scriptsize \textcolor{ForestGreen}{(+1.2)}} & 24.0 {\scriptsize \textcolor{ForestGreen}{(+1.0)}} & 35.4 {\scriptsize \textcolor{ForestGreen}{(+4.3)}} & 18.2 {\scriptsize \textcolor{ForestGreen}{(+1.9)}} & 34.2 \\
    median filtering & 58.6 {\scriptsize \textcolor{ForestGreen}{(+0.5)}} & 23.7 {\scriptsize \textcolor{ForestGreen}{(+0.7)}} & 34.5 {\scriptsize \textcolor{ForestGreen}{(+3.4)}} & 17.6 {\scriptsize \textcolor{ForestGreen}{(+1.3)}} & 33.6 \\
    \bottomrule
    \end{tabular}
}
\caption{{Quantitative evaluation for four suppression strategies (unit: \%).} \textcolor{red}{Red} and \textcolor{ForestGreen}{green} fonts denote descending and ascending values. \textbf{Baseline}: $\attn_{kk}$-proxy CLIP.}
\label{tab:suppressing}
\end{table}
\endgroup
\begin{figure*}[t]
  \includegraphics[width=\textwidth]{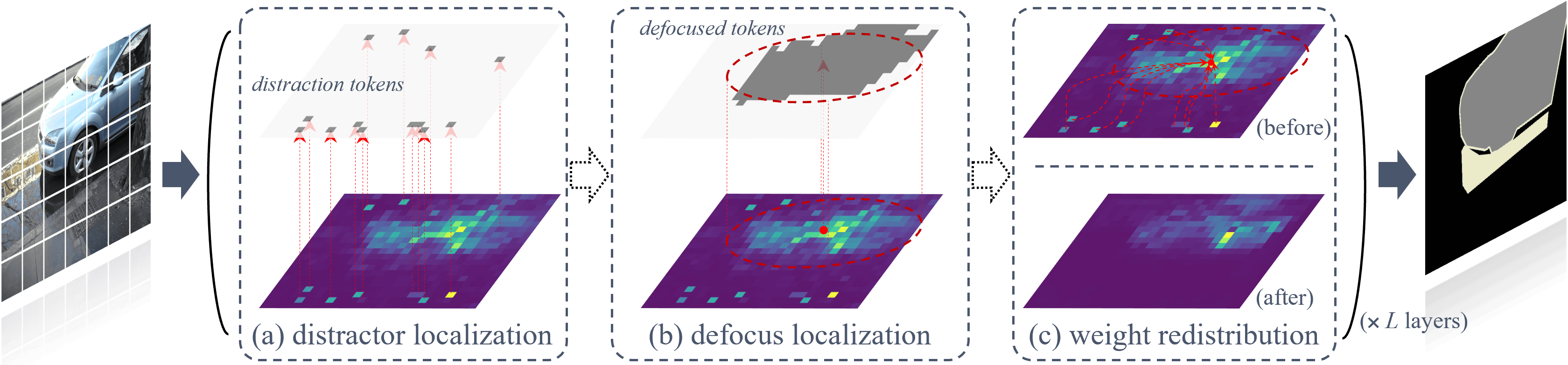}
  \caption{Overview of RF-CLIP. Our RF-CLIP directly correct layer-wise spatial misalignment in CLIP to enhance dense prediction capabilities. Each layer's correction mechanism comprises three key components: (a) distractor localization, identifying attention-rich distraction tokens irrelevant to target objects; (b) defocus localization, detecting attention-poor defocused tokens relevant to targets; (c) weight redistribution, reallocating attention resources from distraction tokens to defocused tokens.}
  \label{fig:overview}
\end{figure*}

\section{Approach} \label{Approach}
Our objective is to improve CLIP’s dense prediction performance by mitigating layer-wise spatial misalignment. Building on the above analysis, we introduce \underline{R}e\underline{F}ocusing CLIP (\ours) – a training-free approach that simulates human distraction-refocusing behavior to redirect layer-wise scattered attention resources toward defocused target regions. As shown in Figure~\ref{fig:overview}, \ours \xspace consists of three key components: (a) Distractor Localization, identifying attention-rich distraction tokens; (b) Defocus Localization, detecting attention-poor defocused tokens; (c) Weight Redistribution, refocusing distracted attention back to defocused target tokens. Below, we use the $l$-th residual layer as an illustrative example to elucidate these components.
\paragraph{Distractor localization.} Given the input embedding $\bm{f}^l_i \in \mathbb{R}^{d}$ of the $i$-th visual token, we define its maximum embedding weight in the distraction dimensions can be defined as: 
\begin{equation}
  \phi^l_i=\max_{j\in \mathcal{D}_{dis}}\frac{\bm{f}^l_i[j]}{\sum^d_{k=1}\bm{f}^l_i[k]}
\end{equation}
Using the threshold $\tau=5/d$, we identify distraction tokens statisfying $\phi^l_i>\tau$.
\paragraph{Defocus localization.} We conceptualize defocused tokens as foreground instances and formulate defocus localization as a bipartition graph cut problem. Adopting spectral clustering on the similarity matrix $\attn^l_{kk}$, we represent each token $i$  as a node $n_i$, partitioning the graph into foreground $\mathcal{A}$ and background $\mathcal{B}$ sets. This optimization problem aims to minimize the normalized cut energy:
\begin{equation}
  \frac{\mathcal{E}(\mathcal{A,B})}{\mathcal{E}(\mathcal{A,V})}+\frac{\mathcal{E}(\mathcal{A,B})}{\mathcal{E}(\mathcal{B,V})}, \quad \mathcal{V}=\mathcal{A}\cup\mathcal{B},
  \label{eq:N-CUT-1}
\end{equation}
where $\mathcal{E}(\cdot ,\cdot)$ denotes set similarity. Using \textit{key-key} attention $\attn^l_{kk}$ as token-wise similarity measure, we define $\mathcal{E}(\mathcal{A,B})=\sum_{n_i\in \mathcal{A},n_j\in \mathcal{B}}\attn^l_{kk}[i,j]$. Following~\citet{shi2000normalized}, Equation~\ref{eq:N-CUT-1} is optimized by solving:
\begin{equation}
  \bm{y}^l_1=\argmin_{{\bm{y}^l}^\top \bm{D1}=0}\frac{{\bm{y}^l}^\top(\bm{D}^l-\attn^l_{kk}){\bm{y}^l}}{{\bm{y}^l}^\top\bm{D}^l{\bm{y}^l}},
  \label{eq:N-CUT-2}
\end{equation}
where $\bm{y}^l_1\in \mathbb{R}^N$ denotes the Fiedler vector — the eigenvector corresponding to the second smallest eigenvalue of the generalized eigensystem $(\bm{D}^l-\attn^l_{kk})\bm{y}^l=\lambda \bm{D}^l \bm{y}^l$. Here, $\bm{D}$ is a diagonal matrix with $\sum_j\attn^l_{kk}[:,j]$ on its diagonal. Thus, defocused tokens are identified as those satisfying $\bm{y}^l_1[i]>\frac{1}{N}\sum^N_{j=1}\bm{y}^l_1[j]$.

\paragraph{Weight redistribution.} We formalize weight redistribution through two complementary mechanisms: (1) attention weight redistribution across spatial locations, and (2) embedding weight redistribution across embedding dimensions. Let $\mathcal{T}_{dis}$ and $\mathcal{T}_{def}$ denote the sets of distraction tokens and defocused tokens, respectively. The attention redistribution process transfers attention weights from $\mathcal{T}_{dis}$ to $\mathcal{T}_{def}$ while preserving original contribution distribution, thereby maintaining the topological structure of layer-wise self-attention matrices and preventing model collapse. This is achieved through two sequential operations. First. attention weights for distraction tokens are scaled down, with the decrement preserved as a redistribution budget $\Omega$:
\begin{align}
  \attn^{l,h}_{qk}[i,j] &\leftarrow (1-\beta) \cdot \attn^{l,h}_{qk}[i,j],_{\,\forall j\in \mathcal{T}_{dis}}, \label{eq:attn-redis-1}\\
  \Omega[i] &=  \beta \cdot \sum_{j\in \mathcal{T}_{dis}}\attn^{l,h}_{qk}[i,j],
\end{align}
where $\beta\in (0,1)$ is the attenuation factor. Second, the budget is distributed to defocused tokens proportional to their original attention weights:
\begin{align}
  \attn^{l,h}_{qk}[i,j] &\leftarrow \attn^{l,h}_{qk}[i,j] + \Omega[i] \cdot \rho[i,j],_{\forall j\in \mathcal{T}_{def}}, \label{eq:attn-redis-2}\\
  \rho[i,j] &= \frac{\attn^{l,h}_{qk}[i,j]}{\sum_{j\in \mathcal{T}_{def}}\attn^{l,h}_{qk}[i,j]},
\end{align}
where $\rho$ represents the original contribution distribution determining each defocused token's allocation share. This process maintains column-normalization ($\sum_j\attn^{l,h}_{qk}[i,j]=1$) and preserves the original attention distribution, thus effectively mitigating model collapse while enhancing focus on defocused tokens. The embedding redistribution process uses a local $3\times 3$ neighborhood centered on each distraction tokens in $\mathcal{T}_{dis}$ to compute averaged embedding representations. For dimension $j\in \mathcal{D}_{dis}$, the embedding weights are updated through spatial averaging:
\begin{align}
  \bm{f}^l_i[j]=\frac{1}{8}\cdot \sum_{\hat{i}\in \mathcal{O}_i}\bm{f}^l_{\hat{i}}[j],_{\forall j\in \mathcal{D}_{dis} \, \text{and}\, i\in \mathcal{T}_{dis}},
  \label{eq:embedding-redis}
\end{align}
where $\mathcal{O}_i$ denote $3\times 3$ neighboring regions centered token $i$. This process only adjusts the embedding in $\mathcal{D}_{dis}$ without destroying the distribution in normal dimensions, thus effectively ensuring the integrity of the original embeddings.
\paragraph{Dense prediction.} Given the layer-wise spatial misalignment correction, we replace $\attn^L_{qk}$ with the layer-averaged attention $\overline\attn_{kk} = \frac{1}{L}\sum^L_{l=1}\attn^l_{kk}$ to produce the final visual embeddings. The segmentation map is computed as:
\begin{equation}
  \bm{M} = \argmax_{N_c} \cos{(\bm{f}^L, \bm{f}_t)},\quad \bm{f}^L = \overline\attn_{kk}\cdot \bm{V}^L.\\
\end{equation}
Please refer to the Appendix for the pseudocode of RF-CLIP.


\begingroup
\renewcommand{\arraystretch}{1.0}
\begin{table*}[t]
  \centering
  \setlength{\tabcolsep}{0.4em}
  \resizebox{\textwidth}{!}{
  \begin{tabular}{lcc|ccccccccc}
    \toprule
    \textbf{{Model}} & {\textbf{Baseline}} & {\textbf{Additional VFM}} & {\textbf{VOC21}} & {\textbf{Context60}} & {\textbf{COCO-Obj}} & {\textbf{VOC20}} & {\textbf{Context59}} & {\textbf{COCO-Stuff}} & {\textbf{Cityscapes}} & {\textbf{ADE20K}} & {\textbf{Avg.}} \\
    \midrule
    \midrule
    \rowcolor{gray!5}
    ReCo~\cite{shin2022reco} & CLIP ViT-B/16  & MaskCLIP & 25.1 & 19.9 & 15.7 & 57.7 & 22.3 & 14.8 & 21.6 & 11.2 & 23.5\\
    \rowcolor{gray!5}
    LaVG~\cite{kang2024defense} & CLIP ViT-B/16  & DINO & {62.1} & 31.6 & 34.2 & {82.5} & 34.7 & 23.2 & 26.2 & 15.8 & 38.8\\
    \rowcolor{gray!5}
    CLIP-DINOiser$^\dagger$~\cite{wysoczanska2024clip} & CLIP ViT-B/16 & DINO & 62.1 & 32.4 & 34.8 & 80.9 & 35.9 & 24.6 & 31.7 & 20.0 & 40.3\\
    \rowcolor{gray!5}
    FreeDa~\cite{barsellotti2024training} & CLIP ViT-B/16 & DINOv2 & - & - & - & 85.6 & \textbf{43.1} & \textbf{27.8} & 36.7 & \textbf{22.4} & -\\
    \rowcolor{gray!5}
    ProxyCLIP$^\ddagger$~\cite{lan2024proxyclip} & CLIP ViT-B/16 & DINO & 59.1 & {35.2} & 36.2 & 78.2 & {38.8} & {26.2} & {38.1} & {19.6} & {41.4}\\
    \rowcolor{gray!5}
    CASS~\cite{kim2025distilling} & CLIP ViT-B/16 & DINO & \underline{65.8} & {36.7} & {37.8} & {\textbf{87.8}} & {40.2} & {26.7} & {39.4} &{20.4} & \underline{44.4} \\
    CLIP~\cite{radford2021learning} & CLIP ViT-B/16  & \ding{55} & 18.6 & 7.8 & 6.5 & 49.1 & 11.2 & 7.2 & 6.7 & 3.2 & 13.8\\
    MaskCLIP~\cite{zhou2022extract} & CLIP ViT-B/16  & \ding{55} & 38.3 & 23.6 & 20.6 & 74.9 & 26.4 & 16.4 & 12.6 & 9.8 & 27.9\\
    GroupViT~\cite{xu2022groupvit} & CLIP ViT-B/16  & \ding{55} & 50.4 & 18.7 & 27.5 & 79.7 & 23.4 & 15.3 & 11.1 & 9.2 & 29.4\\
    CLIPtrase~\cite{shao2024explore} & CLIP ViT-B/16  & \ding{55} & 50.9 & 29.9 & \textbf{43.6} & 81.0 & 33.8 & 22.8 & 21.3 & 16.4 & 32.7\\
    TCL~\cite{cha2023learning} & CLIP ViT-B/16 & \ding{55} & 55.0 & 30.4 & 31.6 & 83.2 & 33.9 & 22.4 & 24.0 & 17.1 & 37.2\\
    CLIPSurgery~\cite{li2023clip} & CLIP ViT-B/16 & \ding{55} & {55.2} & {30.3} & {29.7} & {77.5} & {33.4} & {22.2} & {33.1} &{16.1}&{37.2} \\
    GEM~\cite{bousselham2024grounding} & CLIP ViT-B/16  & \ding{55} & 58.7 & 32.0 & 32.9 & 81.7 & 35.6 & 23.9 & 32.6 & 16.9 & 39.3\\
    CaR~\cite{sun2024clip} & CLIP ViT-B/16 & \ding{55} & 48.6 & 13.6 & 15.4 & 73.7 & 18.4 & - & - & 5.4 & -\\
    ClearCLIP~\cite{lan2024clearclip} & CLIP ViT-B/16 & \ding{55} & 51.8 & 32.6 & 33.0 & 80.9 & 35.9 & 23.9 & 30.0 & 16.7 & 38.1\\
    SCLIP~\cite{wang2024sclip} & CLIP ViT-B/16 & \ding{55} & 59.1 & 30.4 & 30.5 & 80.4 & 34.1 & 22.4 & 32.2 & 16.1 & 38.2\\
    NACLIP~\cite{hajimiri2025pay} & CLIP ViT-B/16 & \ding{55} & 58.9 & 32.2 & 33.2 & 79.7 & 35.2 & 23.3 & 35.5 & 17.4 & 39.4\\
    SC-CLIP~\cite{bai2024self} & CLIP ViT-B/16 & \ding{55} & {64.6} & \underline{36.8} & {37.7} & {84.3} & {40.1} & {26.6} & \underline{41.0} &{20.1} & {43.9} \\
    \rowcolor{blue!5}
    \textbf{\ours \xspace (Ours)}  & CLIP ViT-B/16 & \ding{55} & 64.8 & 36.4 & 37.9 & 87.0 & 39.8 & 26.3 & 41.3 & 20.4 & 44.2\\
    \rowcolor{blue!5}
    \textit{with PAMR}  &  &  & \textbf{67.2} & \textbf{37.9} & \underline{39.1} & \underline{87.0} & \underline{41.4} & \underline{27.5} & \textbf{43.0} & \underline{21.0} & \textbf{45.5}\\
    \midrule
    \rowcolor{gray!5}
    ProxyCLIP$^\ddagger$~\cite{lan2024proxyclip} & CLIP ViT-L/14 & DINO & 58.1 & 34.1 & 37.4 & 82.0 & 37.3 & 25.5 & 38.1 & 21.2 & 41.7\\
    CLIP~\cite{radford2021learning} & CLIP ViT-L/14  & \ding{55} & 10.3 & 4.5 & 4.4 & 19.9 & 5.7 & 3.2 & 3.2 & 1.9 & 6.6\\
    MaskCLIP~\cite{zhou2022extract} & CLIP ViT-L/14  & \ding{55} & 24.8 & 9.7 & 10.2 & 30.1 & 13.0 & 9.0 & 12.1 & 7.1 & 14.5\\
    SCLIP~\cite{wang2024sclip} & CLIP ViT-L/14 & \ding{55} & 44.4 & 22.3 & 24.9 & 70.6 & 25.2 & 16.5 & 21.3 & 10.9 & 29.5\\
    GEM~\cite{bousselham2024grounding} & CLIP ViT-L/14  & \ding{55} & 45.2 & 25.5 & 28.3 & 83.7 & 28.1 & 19.2 & 27.1 & 13.2 & 33.8\\
    CLIPSurgery~\cite{li2023clip} & CLIP ViT-L/14 & \ding{55} & {47.9} & {27.3} & {28.1} & {84.3} & {31.0} & {21.4} & {29.7} &{17.3}&{35.9} \\
    PnP-OVSS~\cite{luo2024emergent} &  CLIP ViT-L/14 & \ding{55} & - & - & 36.2 & 51.3 & 28.0 & 17.9 & - & 14.2 & -\\
    NACLIP~\cite{hajimiri2025pay} & CLIP ViT-L/14 & \ding{55} & 52.1 & 28.7 & 29.9 & 78.6 & 32.1 & 21.4 & 31.4 & 17.3 & 36.4\\
    ClearCLIP~\cite{lan2024clearclip} & CLIP ViT-L/14 & \ding{55} & 48.6 & 28.0 & 28.6 & 84.8 & 31.5 & 21.2 & 32.1 & 16.9 & 36.5\\
    SC-CLIP~\cite{bai2024self} & CLIP ViT-L/14 & \ding{55} & \underline{65.0} & \underline{36.9} & \underline{40.5} & \underline{88.3} & \underline{40.6} & \underline{26.9} & \underline{41.3} & \underline{21.7} & \underline{45.2} \\
    \rowcolor{blue!5}
    \textbf{\ours \xspace (Ours)}  & CLIP ViT-L/14 & \ding{55} & {65.8} & {36.7} & {41.8} & {89.1} & {40.2} & {26.7} & {41.4} &{22.4} &{45.4}\\
    \rowcolor{blue!5}
    \textit{with PAMR}  &  &  & \textbf{68.1} & \textbf{38.1} & \textbf{42.0} & \textbf{89.1} & \textbf{40.8} & \textbf{27.8} & \textbf{43.0} & \textbf{23.4} & \textbf{46.5}\\
\bottomrule
  \end{tabular}
  }
  \caption{{Quantitative evaluation on standard benchmarks (unit: \%).} \textbf{$\dagger$}: Re-implementation with OpenAI's weights. \textbf{$\ddagger$}: Re-implementation with its DINO-B/16 variant. \textbf{\textit{PAMR}}: Pixel-adaptive mask refinement~\cite{araslanov2020single}. Here, the best results are shown in bold and the second-best results are underlined.}
  \label{tab:model_comparison}
\end{table*}
\endgroup
\begingroup
\renewcommand{\arraystretch}{1.0}
\begin{table}[t]
  \centering
  \resizebox{\linewidth}{!}{
  \begin{tabular}{l!{\vrule height 10pt}cccc}
    \toprule
    Model & FLOPs(G)$\downarrow$ & Params(M)$\downarrow$ & Speed(FPS)$\uparrow$  & mIoU(\%)$\uparrow$\\
    \midrule
    \midrule
    baseline & \textbf{16.7}  & 149.6 & \textbf{12.7} & 58.1\\
    CLIP & 17.4  & 149.6 & 12.0 & 18.6\\
    ProxyCLIP & 34.1  & 235.4 & 6.1 & \underline{59.1} \\
    \rowcolor{blue!5}
    \textbf{\ours \xspace (Ours)} & \underline{17.1}  & 149.6 & \underline{12.0} & \textbf{64.8} \\
    \bottomrule
  \end{tabular}
  }
  \caption{{Efficiency analysis on VOC21 benchmark.} \textbf{Baseline}: $\attn_{kk}$-proxy CLIP as the same in Table~\ref{tab:suppressing}.}
  \label{tab:efficiency}
\end{table}
\endgroup
\begingroup
\renewcommand{\arraystretch}{1.0}
\begin{table}[t]
\centering
\resizebox{\linewidth}{!}{
    \begin{tabular}{ll!{\vrule height 10pt}ccccc}
    \toprule
    &Components & VOC21 & COCO-Stuff & Cityscapes & ADE20K & Avg.\\
    \midrule\midrule
    \textbf{(I)} & baseline & 59.1 & 23.6 & 32.1 & 16.9 & 32.9 \\
    \midrule
    \textbf{(II)} & mean filtering   & 58.8 & 21.4 & 31.6 & 14.7 & 31.6\\
    \textbf{(III)} &\textbf{(II)} + Dis Loc.  & 60.3 & 24.4 & 33.6 & 17.5 & 34.0\\
    \textbf{(IV)} &\textbf{(III)} + Attn Red. & 61.5 & 24.8 & 35.3 & 18.3 & 35.0 \\
    \textbf{(V)} &\textbf{(III)} + Embed Red.  & 62.1 & 25.2 & 36.7 & 18.9 & 35.7\\
    \textbf{(VI)} &\textbf{(IV)} + \textbf{(V)}  & 63.2 & 25.4 & 38.5 & 19.3 & 36.6\\
    \rowcolor{blue!5}
    \textbf{(VII)} &\textbf{(VI)} + Def Loc.  & \textbf{64.8} & \textbf{26.3}& \textbf{41.3}& \textbf{20.4}& \textbf{38.2}\\
    \bottomrule
    \end{tabular}
    }
\caption{{Effect of different components (unit: \%).} \textbf{Baseline}: layer-averaged \textit{key-key} attention $\overline\attn_{kk}$-proxy CLIP. \textbf{Dis Loc.}: Distractor localization. \textbf{Attn Red.}: Attention weight redistribution. \textbf{Embed Red.}: Embedding weight redistribution. \textbf{Def Loc.}: Defocus localization.}
\label{tab:component}
\end{table}
\endgroup
\section{Experiments}
\paragraph{Experiments settings.} Following the existing training-free works, we conduct evaluations on eight standard benchmark datasets: VOC21~\cite{pascal-voc-2012}, VOC20~\cite{pascal-voc-2012}, Context60~\cite{mottaghi_cvpr14}, Context59~\cite{mottaghi_cvpr14}, COCO-Suff~\cite{caesar2018cvpr}, COCO-Obj~\cite{caesar2018cvpr}, ADE20K~\cite{zhou2019semantic} and Cityscapes~\cite{cordts2016cityscapes}. Performance is measured using mean \textit{Intersection-over-Union} (mIoU). All experiments are implemented in PyTorch~\cite{paszke2019pytorch} with MMSegmentation~\cite{mmseg2020}. We employ CLIP models with ViT-B/16 and ViT-L/14 architectures, and set $\beta$ to $0.7$. Evaluation configurations include: sliding window inference with a $224\times 224$ window and a $112\times 112$ stride, input resolution resizing with a short side of $336$ ($560$ for Cityscapes), and text descriptions $\{{\bm T}_i\}^{N_c}_{i=1}$ constructed by combining standard ImageNet prompts~\cite{radford2021learning} with category names. See Appendix for more details.
\paragraph{System level comparison.} To evaluate the performance of our proposed method, we compare it with unsupervised approaches, including both training-free and pre-training modes. As shown in Table~\ref{tab:model_comparison}, quantitative evaluation results demonstrate that: 1) Our method achieves the SOTA performance, outperforming same-baseline approaches by $1.6$\% mIoU and even surpassing methods incorporating additional visual foundational models (VFMs) by $+1.1$\% mIoU; 2) While methods like FreeDa~\cite{barsellotti2024training} and CASS~\cite{kim2025distilling} achieve SOTA results on specific benchmarks, they exhibit some performance drops on others. In contrast, our approach maintains top-2 performance across all benchmarks, highlighting superior generalization and robustness; 3) Without utilizing the post-processing module, i.e., PAMR~\cite{araslanov2020single}, our method remains highly competitive and still achieves SOTA performance against same-baseline approaches.
\begin{figure}[t]
  \includegraphics[width=\linewidth]{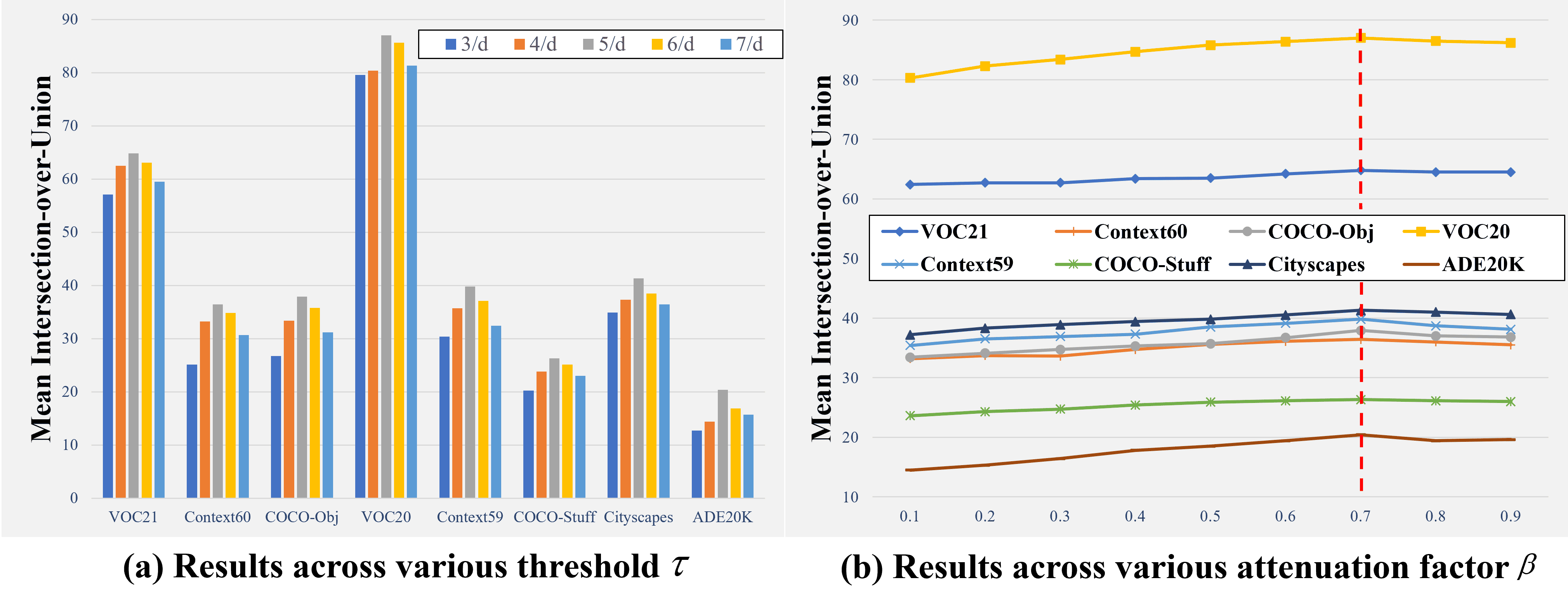}
  \caption{Quantitative results under various threshold and attenuation factor across eight benchmark datasets.}
  \label{fig:threshold}
\end{figure}
\paragraph{Efficiency analysis.} To verify the efficiency of our method, we conduct efficiency analysis on an NVIDIA RTX 3090 GPU. As shown in the Table~\ref{tab:efficiency}, our approach achieves $46.2$\% higher mIoU while maintaining inference speed comparable to CLIP. Compared with ProxyCLIP~\cite{lan2024proxyclip} which introduces an additional VFM, our method doubles the inference speed while improving mIoU by $5.7$\%. In summary, these results demonstrate that our method achieves the optimal performance-efficiency trade-off.
\paragraph{Components analysis.} To validate the effectiveness of the proposed components, we adopt layer-averaged \textit{key-key} attention $\overline\attn_{kk}$-proxy CLIP as the baseline (I), and then incrementally integrate our components to measure performance gains. As shown in Table~\ref{tab:component}, we first perform $3\times 3$ mean filtering on $10$ randomly-selected tokens (II), and then the performance drops by 1.3\% mIoU compared to the baseline. In contrast, performing same operation on distraction tokens improves performance by 1.1\% mIoU (III). These two controlled experiments establish the importance of distraction-aware processing. Next, we reallocate the abundant attention resources from distraction tokens to all non-distraction tokens (IV, V, VI), yielding progressive mIoU gains of $+2.1$\%, $+2.8$\%, and $+3.7$\%. This demonstrates the importance of reallocating the abundant resources occupied by distraction tokens. Finally, defocus localization (VII) directs these liberated resources to target regions, providing an additional $+1.6$\% mIoU improvement. Cumulatively, our components deliver $+1.1$\%, $+2.6$\%, and $+1.6$\% mIoU gains per stage (total $+5.3$\% mIoU), confirming their effectiveness.
\paragraph{Threshold $\tau$ analysis.} The threshold setting directly impacts the selection of distraction tokens. Intuitively, a lower threshold yields more distraction tokens, consequently increasing the false positive rate; conversely, a higher threshold generates fewer distraction tokens, thus elevating the false negative rate. To investigate appropriate threshold configurations, we conducted exploratory experiments with a range of threshold values across eight benchmarks. Figure~\ref{fig:threshold} (a) demonstrates that setting $\tau=5/d$ achieves optimal performance across all benchmarks; moreover, the performance degradation under low thresholds is significantly more pronounced than under high thresholds, indicating a preference for prioritizing the avoidance of false positives (misclassifying normal tokens as distraction tokens) over false negatives (overlooking distraction tokens) in optimization.
\paragraph{Similarity matrix selection on defocus localization.} The similarity matrix assigns weights between nodes and critically influences the outcomes of graph cut. As shown in Table~\ref{tab:ablation_2}, the performance comparison across various matrices indicates that distinct self-attention matrices induce subtle performance variations; however, averaging across all preceding layers generates a more robust similarity measure, leading to enhanced overall performance.
\paragraph{Graph cut threshold analysis.} Table~\ref{tab:ablation_2} presents the performance comparison of two threshold-setting methods: Otsu~\cite{otsu1975threshold} and mean value $\frac{1}{N}\sum^N_{j=1}\bm{y}^l_1[j]$. The results demonstrate that setting the threshold to $\frac{1}{N}\sum^N_{j=1}\bm{y}^l_1[j]$ yields the optimal performance.
\paragraph{Embedding redistribution's receptive field analysis.} In Equation~\ref{eq:embedding-redis}, we employ a $3\times3$ neighborhood. Table~\ref{tab:ablation_2} compares performance across different neighborhood sizes, demonstrating that larger neighborhoods degrade performance. This indicates that distraction tokens are likely concentrated in high-frequency regions, making a smaller receptive field more effective for capturing their characteristics.
\paragraph{Attenuation factor $\beta$ analysis.} The attenuation factor $\beta$ in Equation~\ref{eq:attn-redis-1} governs the allocation of attention resources to defocused tokens; a higher value results in more resources being assigned. Figure~\ref{fig:threshold} (b) displays the quantitative analysis for different $\beta$, indicating that the optimal performance is achieved at $\beta=0.7$.
\begingroup
\renewcommand{\arraystretch}{1.0}
\begin{table}[t]
\centering
\resizebox{\linewidth}{!}{
    \begin{tabular}{ll!{\vrule height 10pt}ccccc}
    \toprule
    & Variants & VOC21 & COCO-Stuff & Cityscapes & ADE20K & Avg.\\
    \midrule
    \midrule
    \multicolumn{7}{c}{similarity matrix selection on defocus localization}\\
    \midrule
    \textbf{(I)} & $\attn^l_{qk}$ & 63.1 & 25.7 & 38.8 & 19.7 & 36.8\\
    \textbf{(II)} & $\attn^l_{qq}$ & 63.5 & 26.0 & 40.2 & 19.1 & 37.2 \\
    \textbf{(III)} & $\attn^l_{kk}$ & 64.0 & 26.1 & 40.9& 19.6 & 37.7 \\
    \rowcolor{blue!5}
    \textbf{(IV)} & $\bar\attn^l_{kk}$ & \textbf{64.8} & \textbf{26.3}& \textbf{41.3}& \textbf{20.4}& \textbf{38.2}\\
    \midrule
    \multicolumn{7}{c}{graph cut threshold analysis}\\
    \midrule
    \textbf{(I)} & Otsu. & 64.5 & 26.3 & 41.1 & 19.5 & 37.9 \\
    \rowcolor{blue!5}
    \textbf{(II)} & Mean.  & \textbf{64.8} & \textbf{26.3}& \textbf{41.3}& \textbf{20.4}& \textbf{38.2}\\
    \midrule
    \multicolumn{7}{c}{embedding redistribution’s receptive field analysis}\\
    \midrule
    \rowcolor{blue!5}
    \textbf{(I)} & $3\times3$ & \textbf{64.8} & \textbf{26.3}& \textbf{41.3}& \textbf{20.4}& \textbf{38.2}\\
    \textbf{(II)} & $5\times5$  & {63.5} & {25.1}& {39.7}& {19.3}&36.9 \\
    \textbf{(III)} & $7\times7$  & 62.2 & 25.0 & 38.4 & 18.7 &36.1 \\
    \bottomrule
    \end{tabular}}
\caption{Effect of different strategy for our components (unit: \%). ${\bar\attn^l_{kk}}$: The averaged $\attn_{kk}$ of all layers preceding the current layer. \textbf{Otsu.}: An automatic threshold selection technique~\cite{otsu1975threshold}. \textbf{Mean.}: Setting the mean value $\frac{1}{N}\sum^N_{j=1}\bm{y}^l_1[j]$ as the threshold.}
\label{tab:ablation_2}
\end{table}
\endgroup

\section{Conclusion}
In this work, we systematically investigate CLIP's internal mechanisms. Our findings reveal that shallow-layer features exhibit strong spatial coherence and high target saliency; however, in deep layers, distraction tokens emerge that consume significant attention resources, thereby disrupting spatial coherence and diverting attention from highly salient target regions. We designate this as the ``distraction'' phenomenon. Further experiments demonstrate that these distraction tokens originate from excessive activation in specific dimensions; based on this insight, we effectively localize distraction tokens and achieve performance improvements by filtering them out. Inspired by this, we propose \ours, a training-free approach that redistributes attention resources from distraction tokens back to target regions. \ours \xspace operates through two stages: first, it identifies distraction tokens and defocused target regions; second, it leverages topology-aware attention reallocation to reassign attention resources occupied by distraction tokens to target regions, thereby mitigating model collapse. Comprehensive experimental evaluations demonstrate significant improvements in dense prediction accuracy and inference speed.

\section*{Acknowledgments}
This work was supported in part by National Natural Science Foundation of China under Grant 62176224, 62176092, 62222602, 62306165, 62172234, 62306165, and 62376233, in part by Science and Technology on Sonar Laboratory under grant 2024-JCJQ-LB-32/07, in part by Xiaomi Young Talents Program award, in part by Natural Science Foundation of Shanghai under Grant 23ZR1420400; and in part by Natural Science Foundation of Chongqing under Grant CSTB2023NSCQ-JQX0007.

\appendix
\section{Details of ``Distraction'' Phenomenon}
This section primarily supplements the details of the ``distraction'' phenomenon. 
\paragraph{Recent works identifying analogous phenomena.} We summarize recent research identifying analogous phenomena: 1) Registers~\cite{darcet2023vision} recognizes and characterizes artifacts in feature maps of supervised and self-supervised Vision Transformer (ViT) networks; these artifacts correspond to high-norm tokens arising during inference, predominantly in background regions with low information content, which are repurposed for internal computations. 2) CLIPtrase~\cite{shao2024explore} and DeCLIP~\cite{wang2025declip} posit that these distraction tokens serve as proxies for the [CLS] token and are misinterpreted by CLIP as global patches. The learning process of the [CLS] token intrinsically amplifies the saliency of global patches. While global patches facilitate [CLS] token learning, they substantially undermine inter-patch semantic correlations. This impairment of semantic correlation proves particularly detrimental for dense feature tasks such as semantic segmentation. Consequently, this likely constitutes a primary reason for CLIP's inherent unsuitability in directly processing dense feature tasks.
\paragraph{The difference in perspective.} We delineate key distinctions between our insights and prior research: 1) We posit that these outlier tokens occupying substantial attention resources stem from CLIP's inherent characteristics—specifically, CLIP's tendency to excessively activate certain tokens in particular channels, a data-agnostic process. Thus, these tokens manifest not merely within background regions; 2) We contend that rather than solely proxying [CLS] tokens for global representation, these tokens primarily serve as proxies for foreground target tokens, a finding evidenced in Table~\ref{tab:CLS}. After reallocating the attention resources dominated by distraction tokens to the [CLS] token, performance exhibits partial improvement relative to the baseline; however, it underperforms compared to assignment to non-distraction tokens. This indicates that the attention resources consumed by distraction tokens originate not solely from the [CLS] token but also from local image patches, e.g., the foreground regions.

\begin{algorithm}[tb]
\caption{\ours \xspace (take the $l$-th layer as an example)}
\label{alg:algorithm}
\textbf{Input}: Visual embeddings $\bm{f}^l\in \mathbb{R}^{N\times d}$ of layer $l$\\
\textbf{Parameter}: Distraction dimension $\mathcal{D}_{dis}$, threshold $\tau$, head number $H$\\
\textbf{Output}: ${\bm{f}}^{l+1}$ after updated
\begin{algorithmic}[1] 
\STATE Calculate maximum embeddings weights for each visual tokens: $\phi^l\in\mathbb{R}^{N}\leftarrow\max_{j\in\mathcal{D}_{dis}}\frac{\bm{f}^l[:, j]}{\sum^d_{k=1}\bm{f}^l[:,k]}$.
\STATE Calculate the averaged \textit{key-key} attention of all layers preceding layer $l$: $\bar\attn^l_{kk}\leftarrow\frac{1}{l\cdot H}\sum^{i,j=l,H}_{i,j=1}\attn^{i,j}_{kk}$.
\STATE Locate distraction tokens: $\mathcal{T}_{dis} \leftarrow (\phi^l>\tau)$. And locate defocused tokens: $\mathcal{T}_{def} \leftarrow (\bm{y}^l_1>\frac{1}{N}\sum^N_{i=1}\bm{y}^l_1[i])$.
\STATE Attention redistribution according to Equation 9 and 11.
\STATE Forward propagation according to Equation 1 and 2.
\STATE Embedding redistribution according to Equation 13 to obtain updated visual embeddings ${\bm{f}}^{l+1}$.
\STATE \textbf{return} ${\bm{f}}^{l+1}$
\end{algorithmic}
\end{algorithm}

\begingroup
\renewcommand{\arraystretch}{1.0}
\begin{table}[t]
\centering
\caption{{Quantitative evaluation for four suppression strategies (unit: \%).} \textbf{Baseline}: $\attn_{kk}$-proxy CLIP. \textbf{To non-distraction tokens}: Allocate the redistribution budget $\Omega$ to non-distraction tokens. \textbf{To [CLS] tokens}: Allocate $\Omega$ to [CLS] tokens. \textbf{To defocused tokens}: Allocate $\Omega$ to defocused tokens.}
\label{tab:CLS}
\resizebox{\linewidth}{!}{
    \begin{tabular}{l!{\vrule height 10pt}cccccc}
    \toprule
    Method & VOC21 & COCO-Stuff & Cityscapes & ADE20k & Avg.\\
    \midrule
    \midrule
    \rowcolor{gray!5}
    baseline & 58.1 & 23.0 & 31.1 & 16.3 & 32.1 \\
    \hdashline
    To non-distraction tokens & 63.2 & 25.4 & 38.5 & 19.3 & 36.6 \\
    To [CLS] token & 61.4 & 23.8 & 34.3 & 17.3 & 34.2 \\
    To defocused tokens & \textbf{64.8} & \textbf{26.3}& \textbf{41.3} & \textbf{20.4} & \textbf{38.2} \\
    \bottomrule
    \end{tabular}
}
\end{table}
\endgroup

\begin{figure}[t]
  \includegraphics[width=\linewidth]{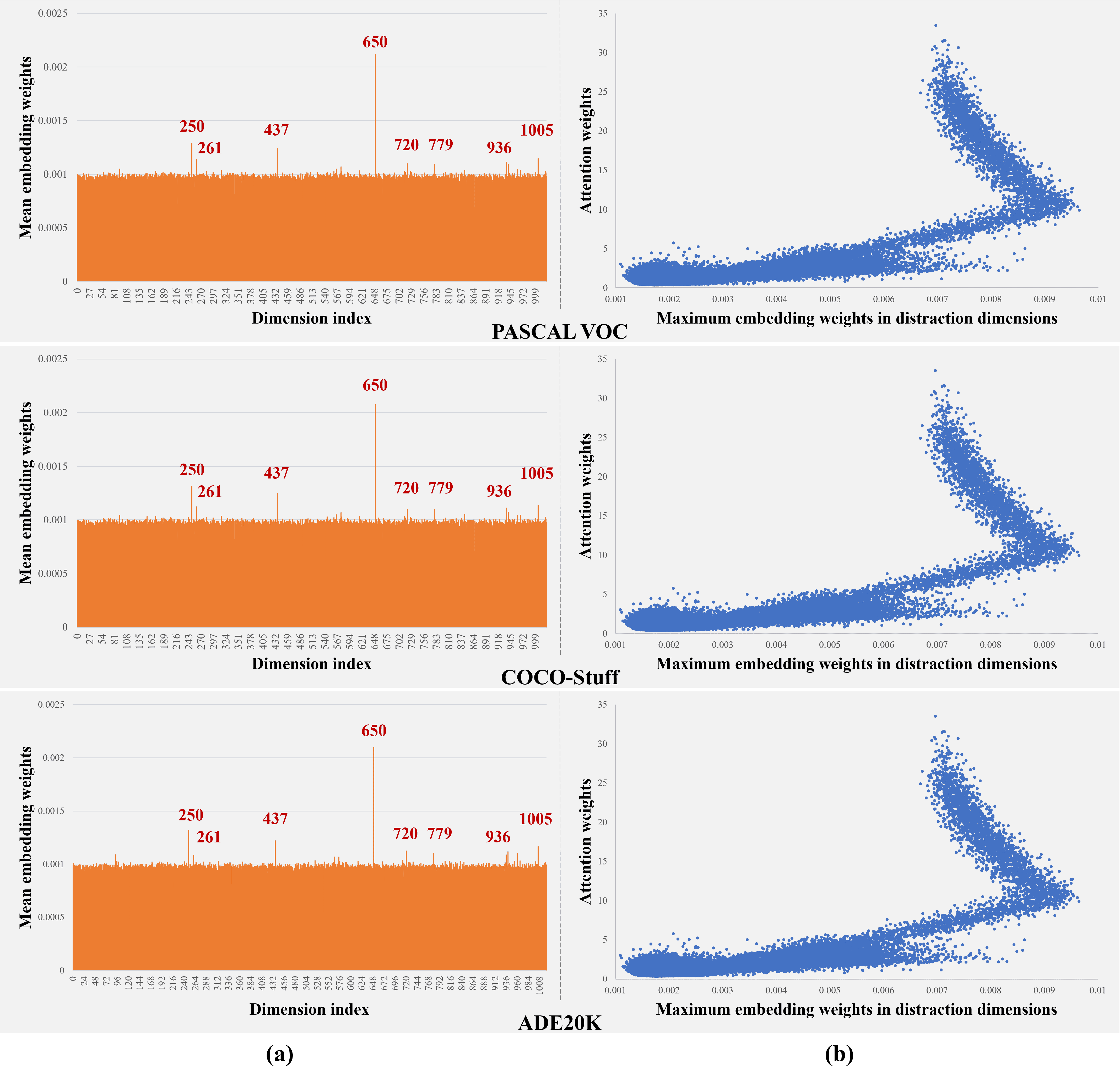}
  \caption{(a) Histogram of per-dimension mean embedding weights averaged across the entire dataset. (b) Scatter plot of all tokens' attention weights versus their maximum embedding weights in distraction dimensions.}
  \label{fig:1}
\end{figure}

\paragraph{Analysis in CLIP-L/14.} In prior sections, we investigated the distribution characteristics of embeddings across dimensions for CLIP-B/16 on three benchmark datasets. In this section, we observe that the same property persists in CLIP-L/14. Specifically, we compute the average vision dense embedding  across all layers using a consistent definition approach: $\bm{\bar f}=\frac{1}{L}\sum^L_{l=1}(\frac{\bm{f}^{l}}{\sum^d_{j=1} \bm{f}^{l}[:,j]}) \in \mathbb{R}^{N\times d}$. Figure~\ref{fig:1} (a) displays the mean embedding weights averaged across the entire dataset, reflecting the weights distribution of the entire dataset in CLIP-L/14‘s high-dimensional space. We find that three datasets exhibit consistent weights distribution, with several peaks at identical dimensions (e.g., 250, 261, 437, etc., denoted red), which we designate as distraction dimensions $\mathcal{D}_{dis}$. This distribution is similar in CLIP-B/16. However, the scatter plot of attention weights versus embedding weights (Figure~\ref{fig:1}(b)) reveals that tokens dominating high attention resources consistently exhibit elevated embedding weights, whereas tokens with high embedding weights do not necessarily possess substantial attention resources. This characteristic diverges from the behavior observed in CLIP-B/16. Consequently, screening distracting tokens solely based on embedding weights would yield high false positives, necessitating joint evaluation of attention weights for accurate identification: $\phi^l_i>\tau \,\land\,\sum_j\attn^l_{qk}[j,i]>15$\footnote{$\tau=6/d$}. We define distraction dimensions: $\mathcal{D}_{dis}=\{4, 162, 189, 326, 429, 474, 633, 713\}$ for CLIP-B/16 and $\mathcal{D}_{dis}=\{250, 261, 437, 650, 720, 779, 936, 1005\}$ for CLIP-L/14.

\begin{figure*}[t]
  \includegraphics[width=\textwidth]{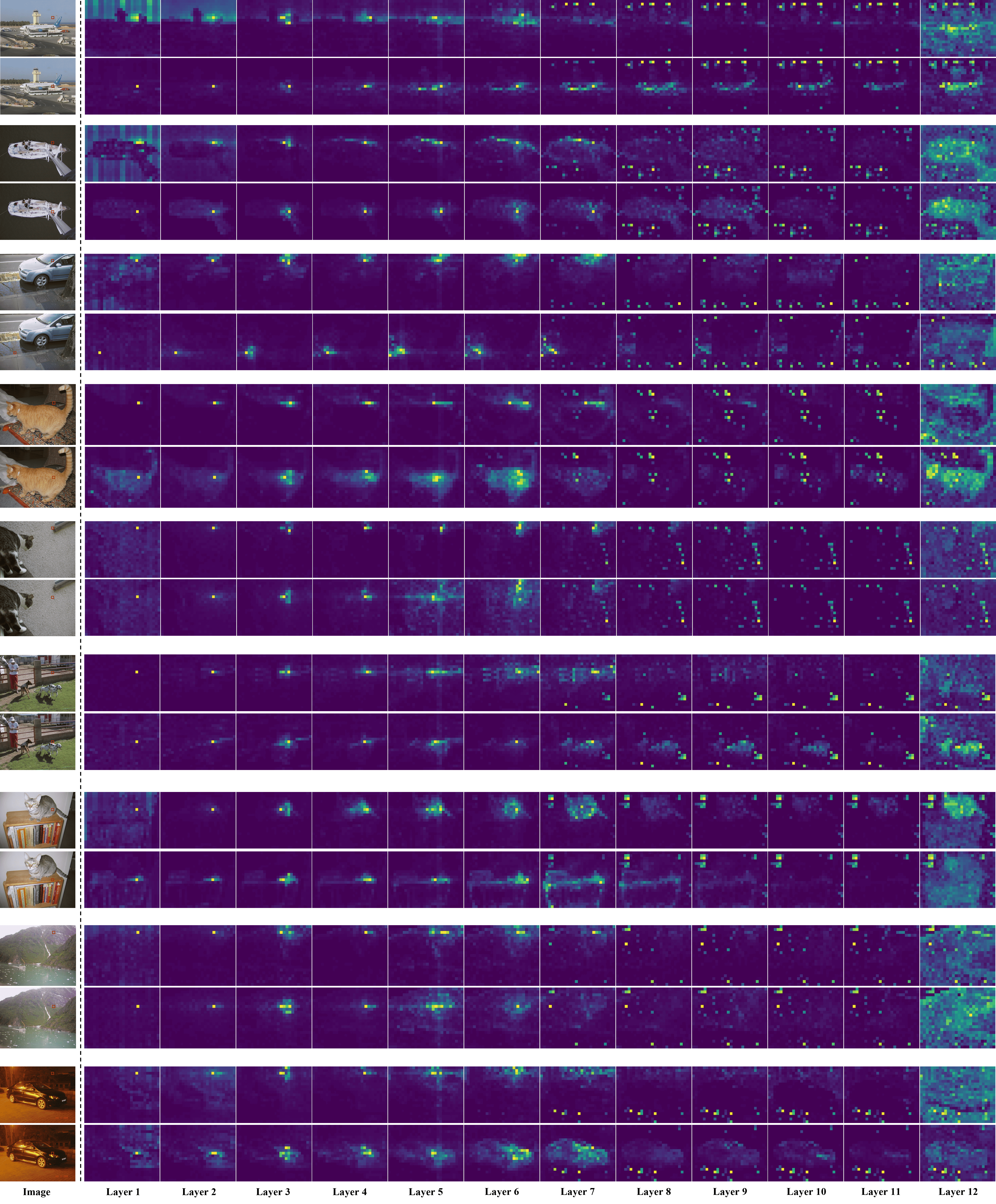}
  \caption{Layer-wise attention maps for two query locations reveal query-to-tokens relevance across the entire image.}
  \label{fig:2}
\end{figure*}


\begin{figure*}[t]
  \includegraphics[width=\textwidth]{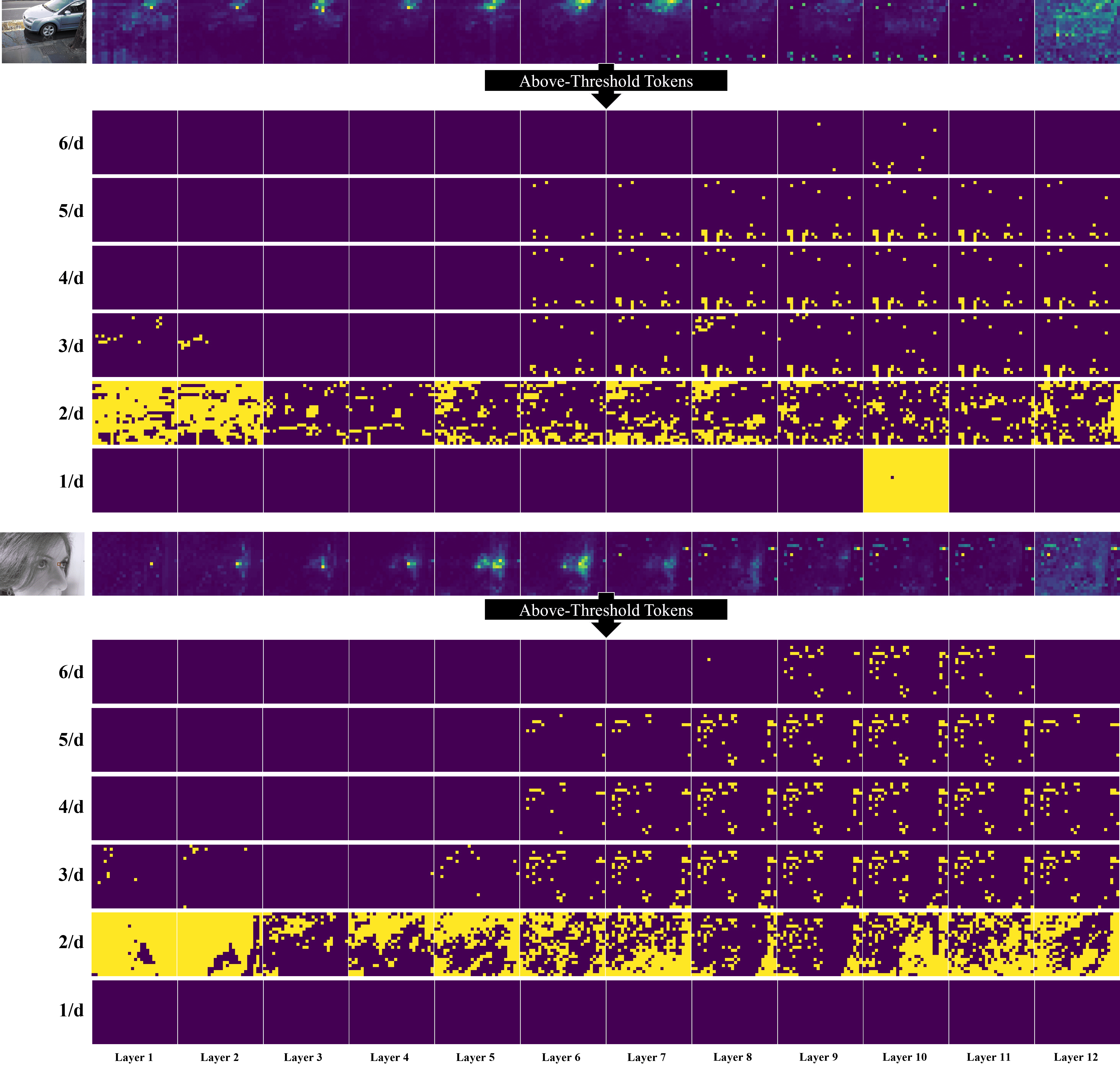}
  \caption{Distraction tokens visualization for various $\tau$.}
  \label{fig:4}
\end{figure*}

\begin{figure*}[t]
  \includegraphics[width=\textwidth]{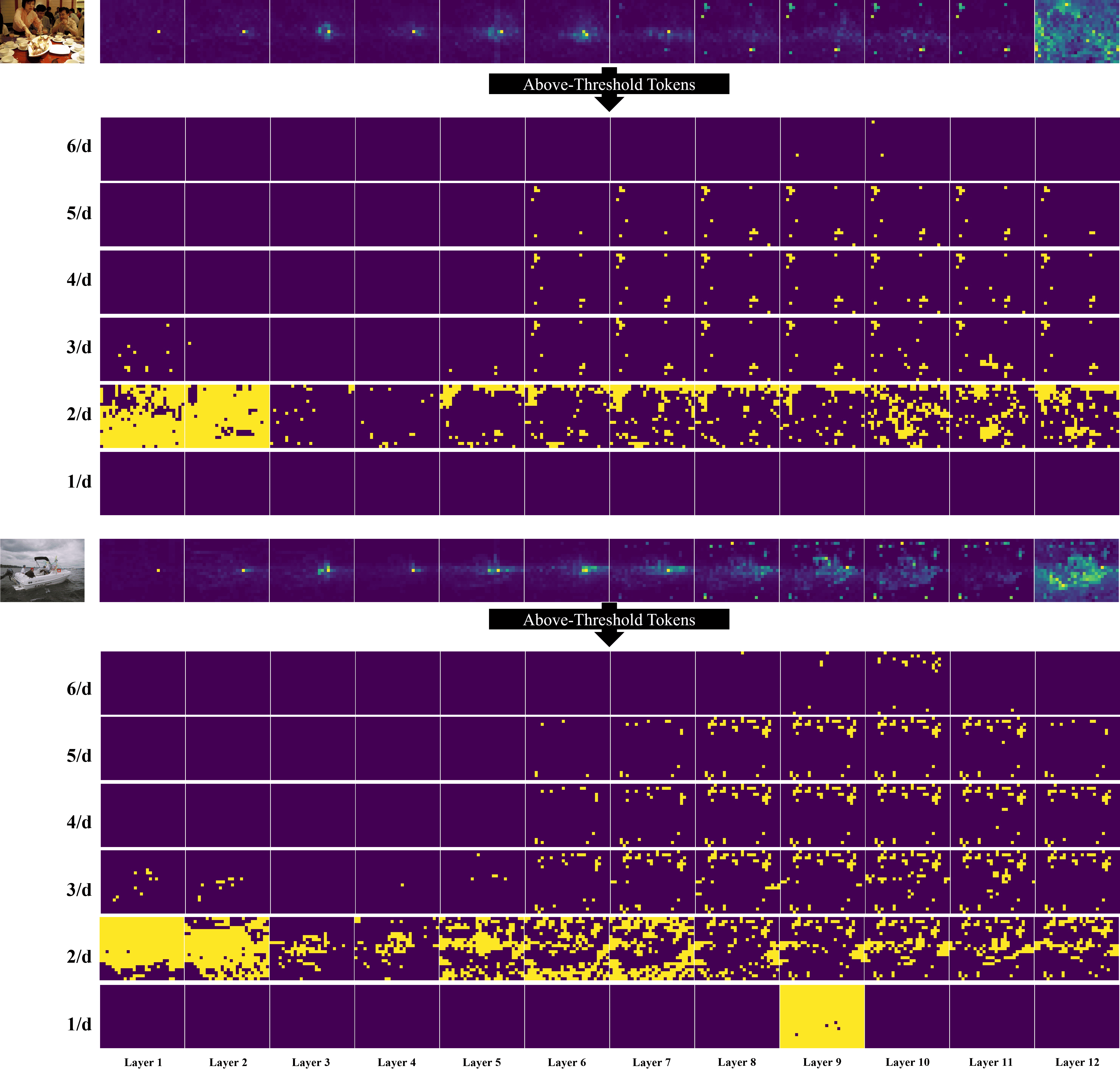}
  \caption{Distraction tokens visualization for various $\tau$.}
  \label{fig:5}
\end{figure*}

\begin{figure*}[t]
  \includegraphics[width=\textwidth]{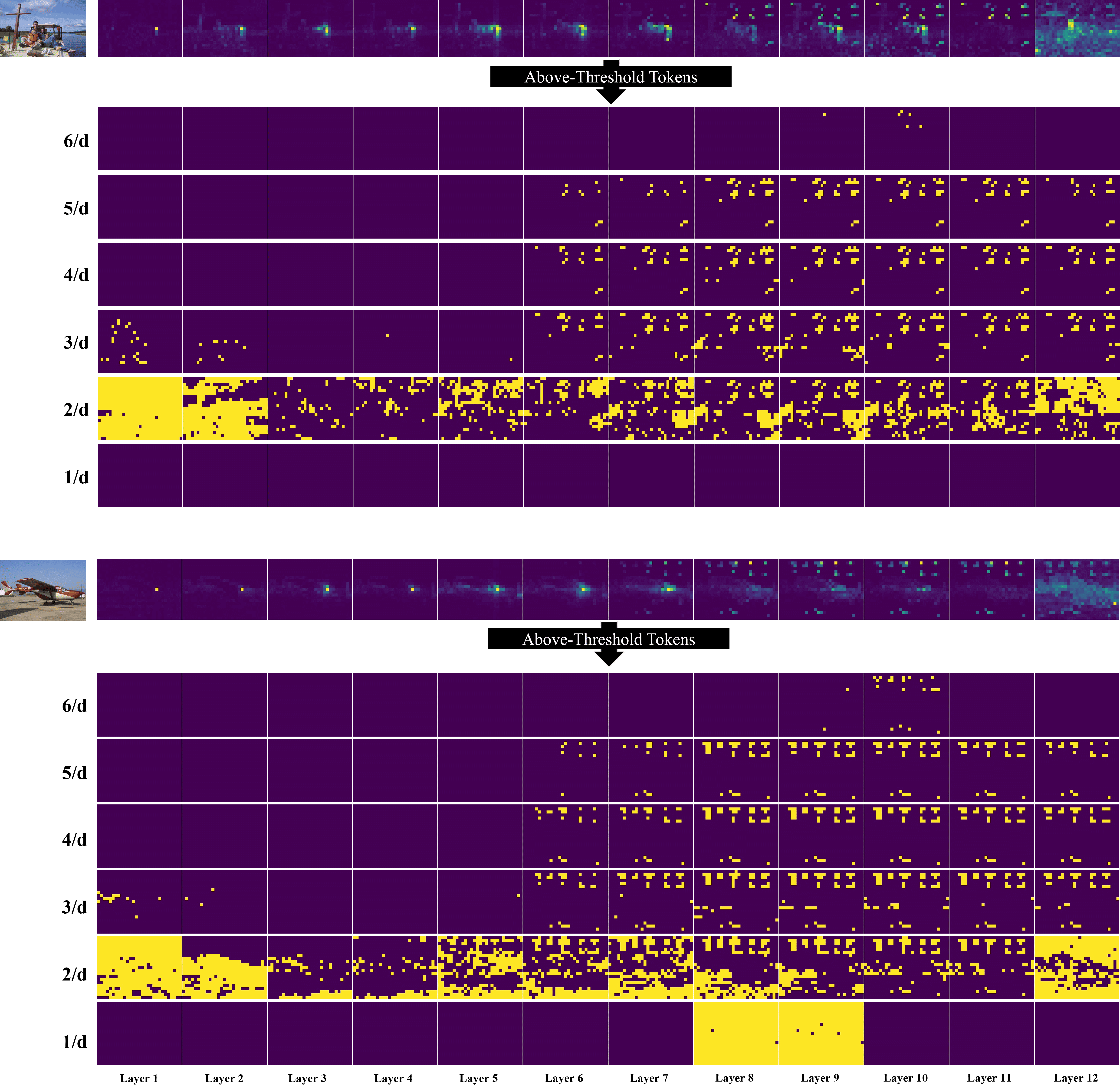}
  \caption{Distraction tokens visualization for various $\tau$.}
  \label{fig:6}
\end{figure*}

\begin{figure*}[t]
  \includegraphics[width=\textwidth]{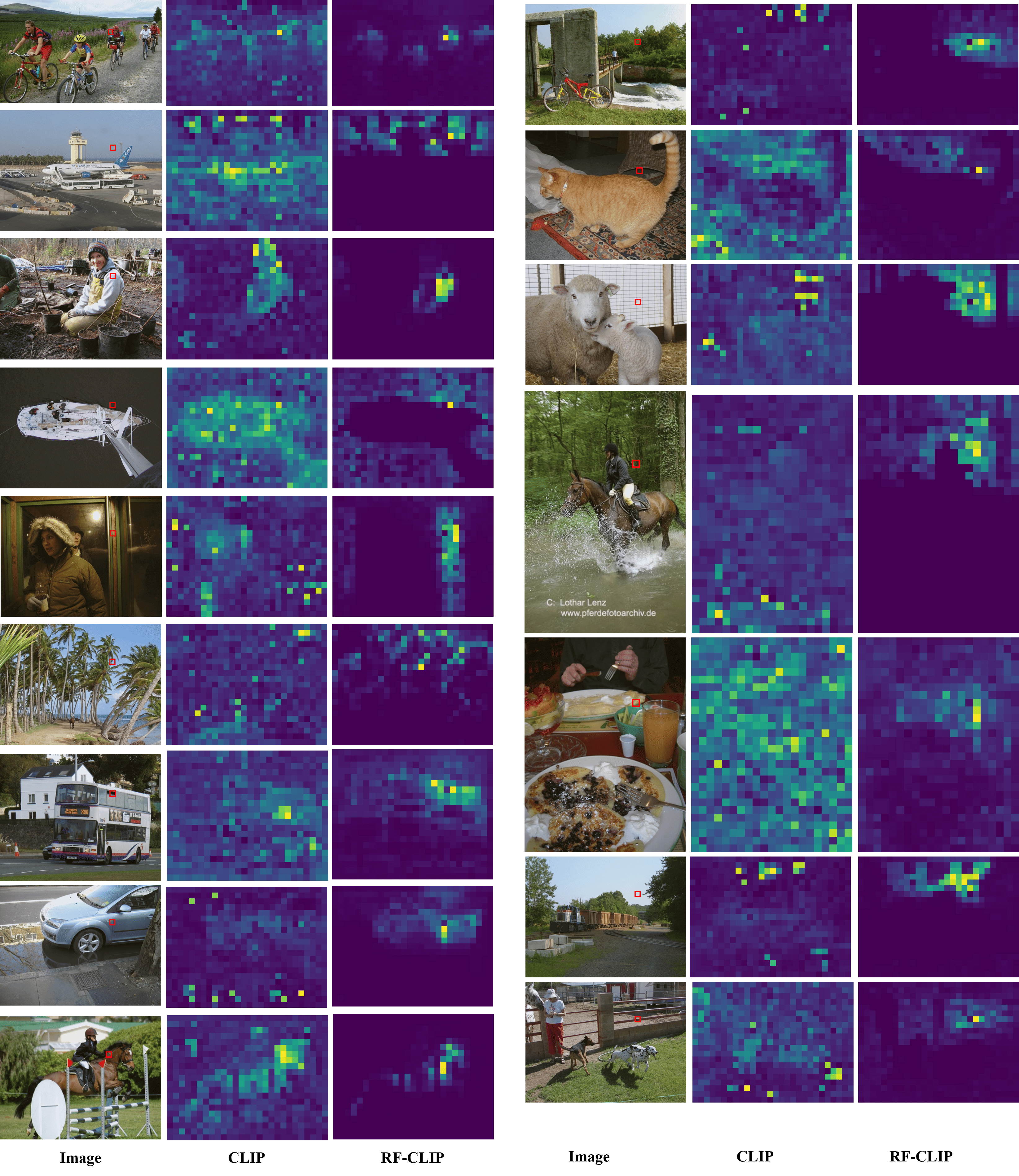}
  \caption{Query-to-tokens relevance visualization.}
  \label{fig:7}
\end{figure*}

\begin{figure*}[t]
  \includegraphics[width=\textwidth]{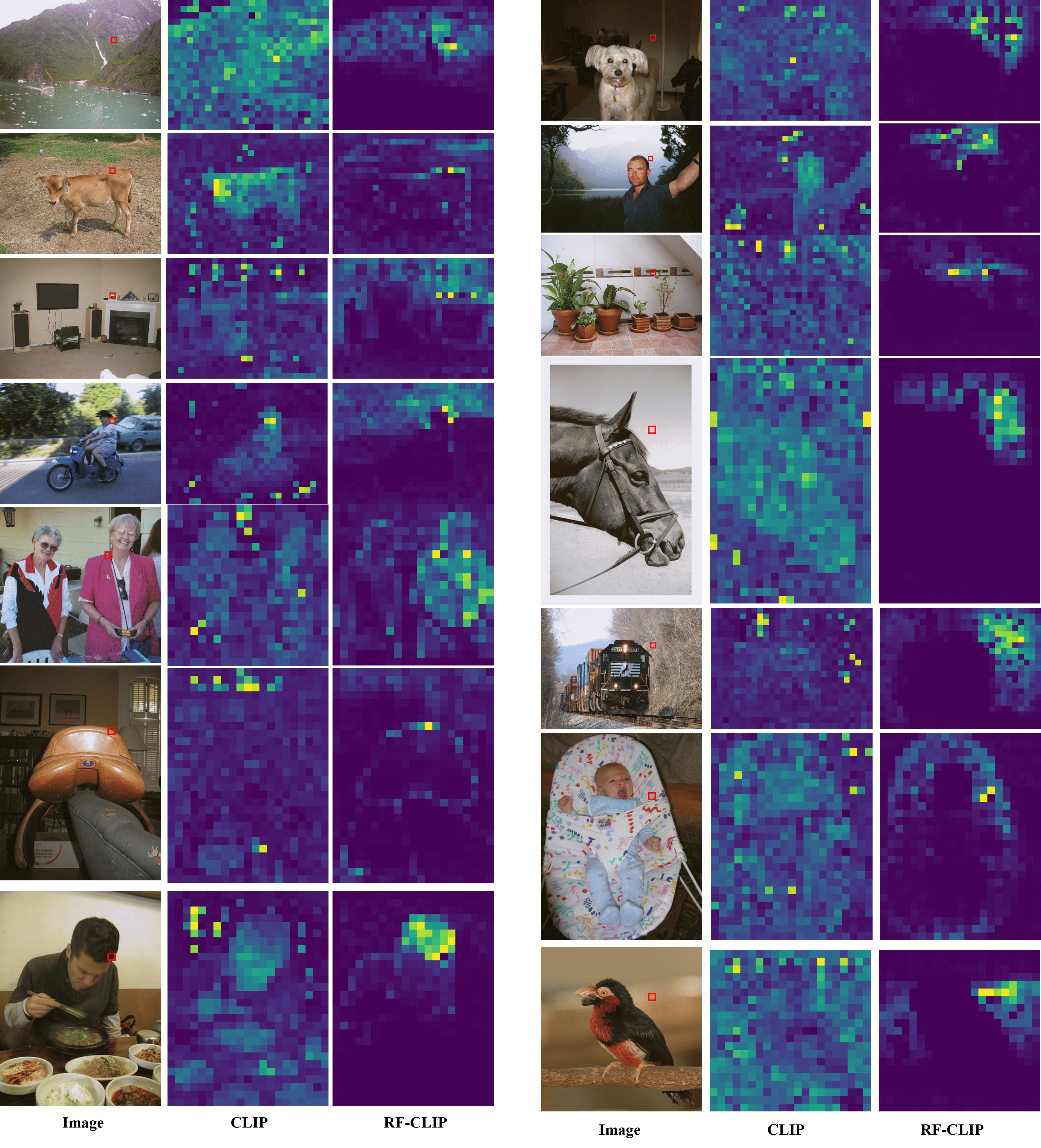}
  \caption{Query-to-tokens relevance visualization.}
  \label{fig:8}
\end{figure*}

\begin{figure*}[t]
  \includegraphics[width=\textwidth]{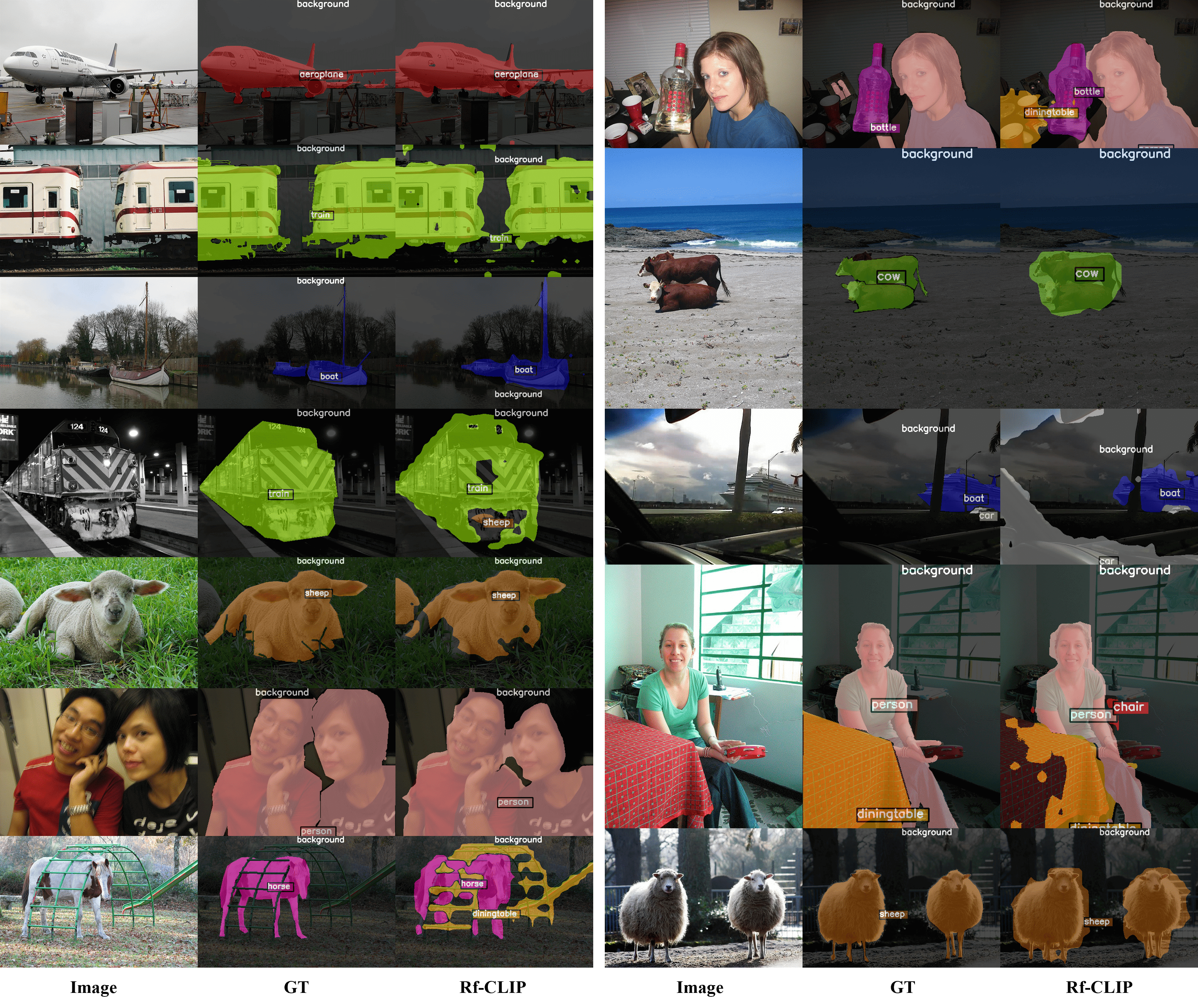}
  \caption{Segmentation visualization on VOC21 benchmark.}
  \label{fig:9}
\end{figure*}

\begin{figure*}[t]
  \includegraphics[width=\textwidth]{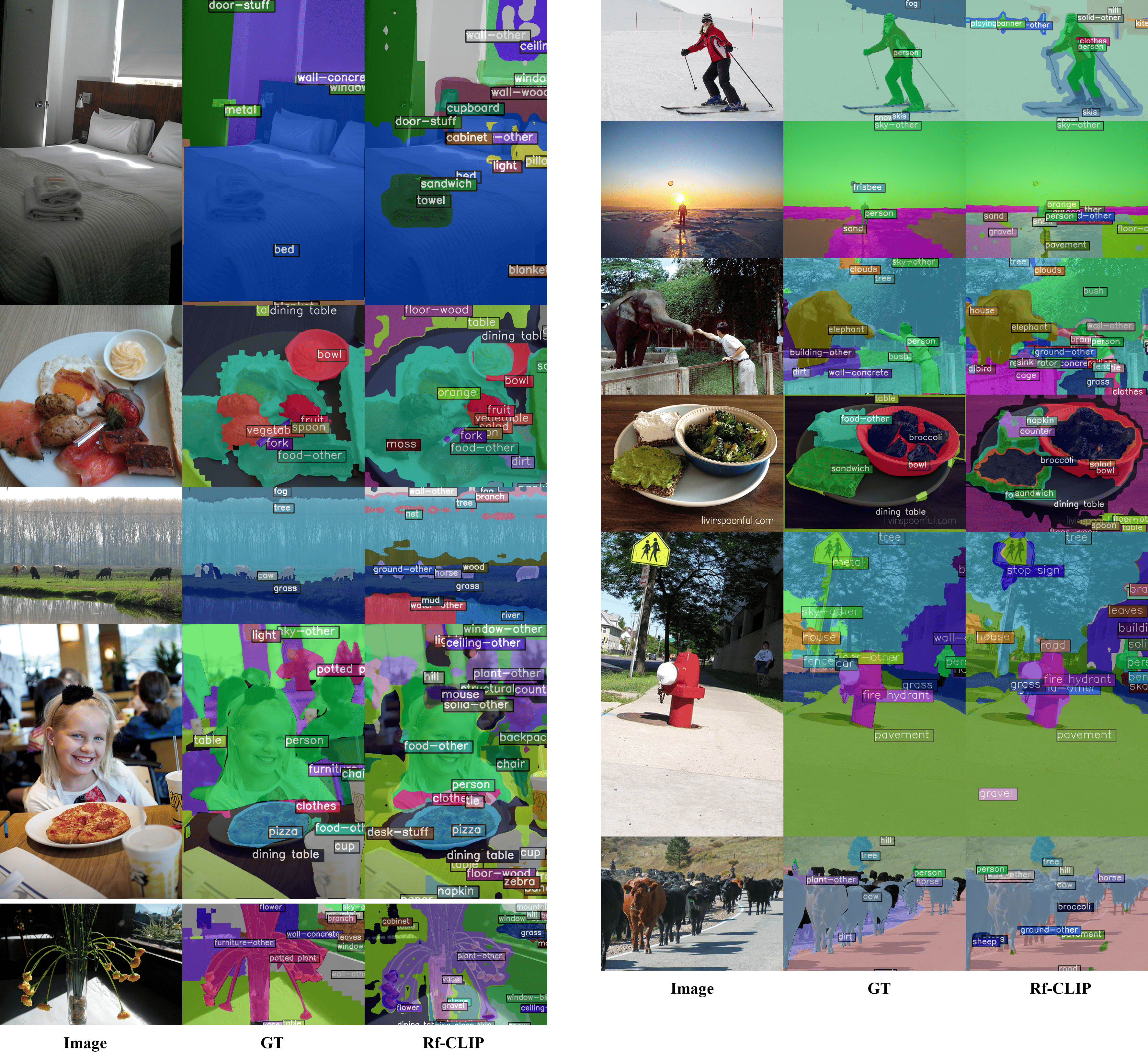}
  \caption{Segmentation visualization on COCO-Stuff benchmark.}
  \label{fig:10}
\end{figure*}



\section{More visualization.} 
This section primarily supplements more visualization analysis for more query-attention(Figure~\ref{fig:2}), distraction tokens (Figure~\ref{fig:4},~\ref{fig:5}, and ~\ref{fig:6}), corrected attention (Figure~\ref{fig:7} and ~\ref{fig:8}), and segmentation results (Figure~\ref{fig:9},~\ref{fig:10}).

\section{More Ablation.} 
\paragraph{Proxy attention analysis.} Our analysis in Table~\ref{tab:2} examines the impact of the final-layer proxy self-attention matrix on model performance. The results demonstrate that the layer-wise averaged \textit{key-key} attention mechanism yields superior performance.
\paragraph{Redistribution layer analysis.} We investigate whether attention reallocation is required for every layer. Table~\ref{tab:2} demonstrates that comprehensive reallocation across all layers yields optimal performance compared to partial layer-wise redistribution strategies.

\begingroup
\renewcommand{\arraystretch}{1.0}
\begin{table}[t]
\centering
\caption{Effect of different strategy for our components (unit: \%). $\overline{\attn}_{qq}$: The averaged $\attn_{qq}$ of all layers. $\overline{\attn}_{kk}$: The averaged $\attn_{kk}$ of all layers.}
\label{tab:2}
\resizebox{\linewidth}{!}{
    \begin{tabular}{ll!{\vrule height 10pt}ccccc}
    \toprule
    & Variants & VOC21 & COCO-Stuff & Cityscapes & ADE20K & Avg.\\
    \midrule
    \midrule
    \multicolumn{7}{c}{last layer's proxy attention selection}\\
    \midrule
    \textbf{(I)} & $\attn^L_{qq}$ & 61.1 & 21.5 & 35.2 & 14.3 & 33.0\\
    \textbf{(II)} & $\attn^L_{kk}$ & 63.3 & 26.1 & 40.2 & 19.5 & 37.3 \\
    \textbf{(III)} & $\overline{\attn}_{qq}$ & 63.7 & 26.4 & 40.6 & 19.2 & 37.5 \\
    \rowcolor{blue!5}
    \textbf{(IV)} & $\overline{\attn}_{kk}$ & \textbf{64.8} & \textbf{26.3}& \textbf{41.3}& \textbf{20.4}& \textbf{38.2}\\
    \midrule
    \multicolumn{7}{c}{redistribution layer analysis}\\
    \midrule
    \textbf{(I)} & $6$ & 64.1 & 25.6 & 41.0 & 19.5 & 37.6 \\
    \textbf{(II)} & $7$ & 64.0 & 25.6 & 40.6 & 19.5 & 37.4 \\
    \textbf{(III)} & $8$ & 64.1 & 25.8 & 40.7 & 19.6 & 37.6 \\
    \textbf{(IV)} & $9$ & 63.9 & 25.3 & 40.5 & 19.8 & 37.6 \\
    \textbf{(V)} & $10$ & 64.1 & 25.8 & 40.8 & 20.1 & 37.9 \\
    \textbf{(VI)} & $11$ & 64.2 & 26.0 & 41.1 & 20.1 & 38.0 \\
    \textbf{(VII)} & $6-12$ & 64.5 & 26.2 & 41.1 & 20.2 & 38.0 \\
    \rowcolor{blue!5}
    \textbf{(VIII)} & $1-12$  & \textbf{64.8} & \textbf{26.3}& \textbf{41.3}& \textbf{20.4}& \textbf{38.2}\\
    \bottomrule
    \end{tabular}}
\end{table}
\endgroup

\section{Related Work}
\paragraph{Open-Vocabulary Semantic Segmentation.} They predominantly dichotomize into two principal paradigms: 1) generative approaches~\cite{gu2020context,bucher2019zero,cheng2021sign,wu2023diffumask,wang2025diffusion} under transductive learning frameworks, and 2) discriminative methods~\cite{ding2022decoupling,zhou2022extract,zhou2023zegclip,xian2019semantic,shi2025llmformer} following inductive learning principles. Generative methods fall under transductive methodologies, typically requiring a prior knowledge about unseen categories in open-world scenarios. These techniques~\cite{xian2019semantic,bucher2019zero,pastore2021closer,gu2020context} synthesize virtual unseen semantic embeddings by amalgamating visual embeddings with textual semantic embeddings from existing priors. Discriminative approaches constitute inductive methodologies that infer unseen semantics through learned categorical knowledge during training, thereby eliminating the need for prior knowledge about novel categories. State-of-the-art implementations predominantly employ either knowledge distillation~\cite{ding2022decoupling,xu2022simple,yu2023zero,liang2023open} or feature adaptation strategies~\cite{zhou2023zegclip,kwon2023probabilistic,xu2023side,xie2023sed,cho2023cat,li2024relationship}. Knowledge distillation methods typically combine VLM-derived image-level semantic discriminability with mask-aware segmentation networks to achieve open-vocabulary segmentation, while feature adaptation approaches directly fine-tune VLMs as backbone networks to convert image-level classification capabilities into pixel-level discriminative power. Despite remarkable progress, current methodologies exhibit insufficient exploration of latent semantic perception mechanisms. 
\paragraph{Joint Fine-tuning in OVSS.} These methods fine-tune CLIP while training an additional decoder. For instance, CAT-Seg~\cite{cho2023cat} proposes an cost-based fine-tuning strategy based on CLIP. MAFT~\cite{jiao2023learning} leverages attention bias to finetune CLIP for classification.
\paragraph{Pre Fine-tuning in OVSS.} These methods directly fine-tune CLIP using fine-grained vision-language contrastive learning.For example, CLIM~\cite{wu2024clim} employs mosaic augmentation to composite multiple images into a single frame. This arrangement enables each sub-image to serve as a pseudo-region for region-text contrastive learning. Similarly, CLIPSelf~\cite{wu2023clipself} enhances CLIP's region-level classification accuracy by maximizing the cosine similarity between the region representations it produces and the representations of their corresponding crops.
\paragraph{Visual Tuning in Vision-Language Models} VLM for vision tasks~\cite{radford2021learning,cui2022contrastive,wu2021data,luddecke2022image,ppl,regzsl,unidseg,ftd} are optimized with a large scale of image-text pair data on the internet. There are three categories: contrastive, generative, and aligned objectives. CLIP ~\cite{radford2021learning} first proposes the paradigm of pre-trained vision-language model. DeCLIP~\cite{li2021supervision} argues that CLIP is data-intensive and proposes a data-efficient training paradigm. UniCL~\cite{yang2022unified} combines the two data sources to build a new image-text-label field and proposes unified contrastive learning. ZeroVL~\cite{cui2022contrastive} proposes debiased sampling to deal with biased representation distributions and a new mixup method for the image and text models. OTTER~\cite{wu2021data} uses optimal transport to find the soft label for contrastive learning and handle the problem of noisy image-text pairs. Visual Prompt Learning~\cite{zhou2022conditional,zang2022unified,khattak2023maple} is a technique that assists in adapting CLIP-like vision-language models for various visual tasks.  CoOp~\cite{zhou2022learning} adopts trainable vectors as word prompt to adapt CLIP for vision classification. VP~\cite{bahng2022exploring} utilizes perturbations as visual prompt. VPT~\cite{jia2022visual} proposes trainable visual prompt to adapt each layer of the visual embeddings. UPT~\cite{zang2022unified} constructs unified prompt modeling to extract trainable visual and textual prompt for adapting CLIP. MaPLe~\cite{khattak2023maple} adopts trainable prompt to guide both visual and textual embeddings and proposes a coupling function as a bridge to build a multi-modal prompt. DenseCLIP~\cite{rao2022denseclip} uses the contextual information from the image to prompt the language model. Probabilistic prompt~\cite{kwon2023probabilistic} applies multiple prompt sampled from probabilistic text embeddings to better understand the image. SegPrompt~\cite{zhu2023segprompt} proposes a category-level prompt to improve the model’s class-agnostic segmentation ability.

\bibliography{main}

@inproceedings{lan2024clearclip,
  title={Clearclip: Decomposing clip representations for dense vision-language inference},
  author={Lan, Mengcheng and Chen, Chaofeng and Ke, Yiping and Wang, Xinjiang and Feng, Litong and Zhang, Wayne},
  booktitle={European Conference on Computer Vision},
  pages={143--160},
  year={2024},
  organization={Springer}
}

@article{shi2000normalized,
  title={Normalized cuts and image segmentation},
  author={Shi, Jianbo and Malik, Jitendra},
  journal={IEEE Transactions on pattern analysis and machine intelligence},
  volume={22},
  number={8},
  pages={888--905},
  year={2000},
  publisher={Ieee}
}

@article{paszke2019pytorch,
  title={Pytorch: An imperative style, high-performance deep learning library},
  author={Paszke, A},
  journal={arXiv preprint arXiv:1912.01703},
  year={2019}
}

@inproceedings{radford2021learning,
  title={Learning transferable visual models from natural language supervision},
  author={Radford, Alec and Kim, Jong Wook and Hallacy, Chris and Ramesh, Aditya and Goh, Gabriel and Agarwal, Sandhini and Sastry, Girish and Askell, Amanda and Mishkin, Pamela and Clark, Jack and others},
  booktitle={International conference on machine learning},
  pages={8748--8763},
  year={2021},
  organization={PmLR}
}

@misc{pascal-voc-2012,
	author = "Everingham, M. and Van~Gool, L. and Williams, C. K. I. and Winn, J. and Zisserman, A.",
	title = "The {PASCAL} {V}isual {O}bject {C}lasses {C}hallenge 2012 {(VOC2012)} {R}esults", 
        year={2012}}

@InProceedings{mottaghi_cvpr14,
 author       = {Roozbeh Mottaghi and Xianjie Chen and Xiaobai Liu and Nam-Gyu Cho and Seong-Whan Lee and Sanja Fidler and Raquel Urtasun and Alan Yuille},
 title        = {The Role of Context for Object Detection and Semantic Segmentation in the Wild},
 booktitle    = {IEEE Conference on Computer Vision and Pattern Recognition (CVPR)},
 year         = {2014},
}

@INPROCEEDINGS{caesar2018cvpr,
  title={COCO-Stuff: Thing and stuff classes in context},
  author={Caesar, Holger and Uijlings, Jasper and Ferrari, Vittorio},
  booktitle={Computer Vision and Pattern Recognition (CVPR), 2018 IEEE conference on},
  organization={IEEE},
  year={2018}
}

@article{zhou2019semantic,
  title={Semantic understanding of scenes through the ade20k dataset},
  author={Zhou, Bolei and Zhao, Hang and Puig, Xavier and Xiao, Tete and Fidler, Sanja and Barriuso, Adela and Torralba, Antonio},
  journal={International Journal of Computer Vision},
  volume={127},
  pages={302--321},
  year={2019},
  publisher={Springer}
}

@inproceedings{cordts2016cityscapes,
  title={The cityscapes dataset for semantic urban scene understanding},
  author={Cordts, Marius and Omran, Mohamed and Ramos, Sebastian and Rehfeld, Timo and Enzweiler, Markus and Benenson, Rodrigo and Franke, Uwe and Roth, Stefan and Schiele, Bernt},
  booktitle={Proceedings of the IEEE conference on computer vision and pattern recognition},
  pages={3213--3223},
  year={2016}
}

@misc{mmseg2020,
    title={{MMSegmentation}: OpenMMLab Semantic Segmentation Toolbox and Benchmark},
    author={MMSegmentation Contributors},
    howpublished = {\url{https://github.com/open-mmlab/mmsegmentation}},
    year={2020}
}

@inproceedings{xu2022groupvit,
  title={Groupvit: Semantic segmentation emerges from text supervision},
  author={Xu, Jiarui and De Mello, Shalini and Liu, Sifei and Byeon, Wonmin and Breuel, Thomas and Kautz, Jan and Wang, Xiaolong},
  booktitle={Proceedings of the IEEE/CVF conference on computer vision and pattern recognition},
  pages={18134--18144},
  year={2022}
}

@inproceedings{cha2023learning,
  title={Learning to generate text-grounded mask for open-world semantic segmentation from only image-text pairs},
  author={Cha, Junbum and Mun, Jonghwan and Roh, Byungseok},
  booktitle={Proceedings of the IEEE/CVF Conference on Computer Vision and Pattern Recognition},
  pages={11165--11174},
  year={2023}
}

@inproceedings{wysoczanska2024clip,
  title={CLIP-DINOiser: Teaching CLIP a few DINO tricks for open-vocabulary semantic segmentation},
  author={Wysocza{\'n}ska, Monika and Sim{\'e}oni, Oriane and Ramamonjisoa, Micha{\"e}l and Bursuc, Andrei and Trzci{\'n}ski, Tomasz and P{\'e}rez, Patrick},
  booktitle={European Conference on Computer Vision},
  pages={320--337},
  year={2024},
  organization={Springer}
}

@article{shin2022reco,
  title={Reco: Retrieve and co-segment for zero-shot transfer},
  author={Shin, Gyungin and Xie, Weidi and Albanie, Samuel},
  journal={Advances in Neural Information Processing Systems},
  volume={35},
  pages={33754--33767},
  year={2022}
}

@inproceedings{barsellotti2024training,
  title={Training-free open-vocabulary segmentation with offline diffusion-augmented prototype generation},
  author={Barsellotti, Luca and Amoroso, Roberto and Cornia, Marcella and Baraldi, Lorenzo and Cucchiara, Rita},
  booktitle={Proceedings of the IEEE/CVF Conference on Computer Vision and Pattern Recognition},
  pages={3689--3698},
  year={2024}
}

@inproceedings{zhou2022extract,
  title={Extract free dense labels from clip},
  author={Zhou, Chong and Loy, Chen Change and Dai, Bo},
  booktitle={European Conference on Computer Vision},
  pages={696--712},
  year={2022},
  organization={Springer}
}

@inproceedings{bousselham2024grounding,
  title={Grounding everything: Emerging localization properties in vision-language transformers},
  author={Bousselham, Walid and Petersen, Felix and Ferrari, Vittorio and Kuehne, Hilde},
  booktitle={Proceedings of the IEEE/CVF Conference on Computer Vision and Pattern Recognition},
  pages={3828--3837},
  year={2024}
}

@inproceedings{sun2024clip,
  title={Clip as rnn: Segment countless visual concepts without training endeavor},
  author={Sun, Shuyang and Li, Runjia and Torr, Philip and Gu, Xiuye and Li, Siyang},
  booktitle={Proceedings of the IEEE/CVF Conference on Computer Vision and Pattern Recognition},
  pages={13171--13182},
  year={2024}
}

@inproceedings{luo2024emergent,
  title={Emergent open-vocabulary semantic segmentation from off-the-shelf vision-language models},
  author={Luo, Jiayun and Khandelwal, Siddhesh and Sigal, Leonid and Li, Boyang},
  booktitle={Proceedings of the IEEE/CVF Conference on Computer Vision and Pattern Recognition},
  pages={4029--4040},
  year={2024}
}

@inproceedings{shao2024explore,
  title={Explore the potential of clip for training-free open vocabulary semantic segmentation},
  author={Shao, Tong and Tian, Zhuotao and Zhao, Hang and Su, Jingyong},
  booktitle={European Conference on Computer Vision},
  pages={139--156},
  year={2024},
  organization={Springer}
}

@inproceedings{wang2024sclip,
  title={Sclip: Rethinking self-attention for dense vision-language inference},
  author={Wang, Feng and Mei, Jieru and Yuille, Alan},
  booktitle={European Conference on Computer Vision},
  pages={315--332},
  year={2024},
  organization={Springer}
}

@inproceedings{kang2024defense,
  title={In defense of lazy visual grounding for open-vocabulary semantic segmentation},
  author={Kang, Dahyun and Cho, Minsu},
  booktitle={European Conference on Computer Vision},
  pages={143--164},
  year={2024},
  organization={Springer}
}

@inproceedings{lan2024proxyclip,
  title={Proxyclip: Proxy attention improves clip for open-vocabulary segmentation},
  author={Lan, Mengcheng and Chen, Chaofeng and Ke, Yiping and Wang, Xinjiang and Feng, Litong and Zhang, Wayne},
  booktitle={European Conference on Computer Vision},
  pages={70--88},
  year={2024},
  organization={Springer}
}

@inproceedings{hajimiri2025pay,
  title={Pay attention to your neighbours: Training-free open-vocabulary semantic segmentation},
  author={Hajimiri, Sina and Ayed, Ismail Ben and Dolz, Jose},
  booktitle={2025 IEEE/CVF Winter Conference on Applications of Computer Vision (WACV)},
  pages={5061--5071},
  year={2025},
  organization={IEEE}
}

@inproceedings{kim2025distilling,
  title={Distilling spectral graph for object-context aware open-vocabulary semantic segmentation},
  author={Kim, Chanyoung and Ju, Dayun and Han, Woojung and Yang, Ming-Hsuan and Hwang, Seong Jae},
  booktitle={Proceedings of the Computer Vision and Pattern Recognition Conference},
  pages={15033--15042},
  year={2025}
}

@article{bai2024self,
  title={Self-calibrated clip for training-free open-vocabulary segmentation},
  author={Bai, Sule and Liu, Yong and Han, Yifei and Zhang, Haoji and Tang, Yansong},
  journal={arXiv preprint arXiv:2411.15869},
  year={2024}
}

@article{li2023clip,
  title={Clip surgery for better explainability with enhancement in open-vocabulary tasks},
  author={Li, Yi and Wang, Hualiang and Duan, Yiqun and Li, Xiaomeng},
  journal={arXiv e-prints},
  pages={arXiv--2304},
  year={2023}
}

@article{zhang2022dino,
  title={Dino: Detr with improved denoising anchor boxes for end-to-end object detection},
  author={Zhang, Hao and Li, Feng and Liu, Shilong and Zhang, Lei and Su, Hang and Zhu, Jun and Ni, Lionel M and Shum, Heung-Yeung},
  journal={arXiv preprint arXiv:2203.03605},
  year={2022}
}

@article{oquab2023dinov2,
  title={Dinov2: Learning robust visual features without supervision},
  author={Oquab, Maxime and Darcet, Timoth{\'e}e and Moutakanni, Th{\'e}o and Vo, Huy and Szafraniec, Marc and Khalidov, Vasil and Fernandez, Pierre and Haziza, Daniel and Massa, Francisco and El-Nouby, Alaaeldin and others},
  journal={arXiv preprint arXiv:2304.07193},
  year={2023}
}

@inproceedings{araslanov2020single,
  title={Single-stage semantic segmentation from image labels},
  author={Araslanov, Nikita and Roth, Stefan},
  booktitle={Proceedings of the IEEE/CVF conference on computer vision and pattern recognition},
  pages={4253--4262},
  year={2020}
}

@article{otsu1975threshold,
  title={A threshold selection method from gray-level histograms},
  author={Otsu, Nobuyuki and others},
  journal={Automatica},
  volume={11},
  number={285-296},
  pages={23--27},
  year={1975}
}

@inproceedings{cho2024cat,
  title={Cat-seg: Cost aggregation for open-vocabulary semantic segmentation},
  author={Cho, Seokju and Shin, Heeseong and Hong, Sunghwan and Arnab, Anurag and Seo, Paul Hongsuck and Kim, Seungryong},
  booktitle={Proceedings of the IEEE/CVF Conference on Computer Vision and Pattern Recognition},
  pages={4113--4123},
  year={2024}
}

@article{jiao2023learning,
  title={Learning mask-aware clip representations for zero-shot segmentation},
  author={Jiao, Siyu and Wei, Yunchao and Wang, Yaowei and Zhao, Yao and Shi, Humphrey},
  journal={Advances in Neural Information Processing Systems},
  volume={36},
  pages={35631--35653},
  year={2023}
}

@article{li2024relationship,
  title={Relationship prompt learning is enough for open-vocabulary semantic segmentation},
  author={Li, Jiahao and Lu, Yang and Xie, Yuan and Qu, Yanyun},
  journal={Advances in Neural Information Processing Systems},
  volume={37},
  pages={74298--74324},
  year={2024}
}

@article{wu2023clipself,
  title={Clipself: Vision transformer distills itself for open-vocabulary dense prediction},
  author={Wu, Size and Zhang, Wenwei and Xu, Lumin and Jin, Sheng and Li, Xiangtai and Liu, Wentao and Loy, Chen Change},
  journal={arXiv preprint arXiv:2310.01403},
  year={2023}
}

@inproceedings{wu2024clim,
  title={Clim: Contrastive language-image mosaic for region representation},
  author={Wu, Size and Zhang, Wenwei and Xu, Lumin and Jin, Sheng and Liu, Wentao and Loy, Chen Change},
  booktitle={Proceedings of the AAAI Conference on Artificial Intelligence},
  volume={38},
  number={6},
  pages={6117--6125},
  year={2024}
}

@article{shi2025llmformer,
  title={LLMFormer: Large language model for open-vocabulary semantic segmentation},
  author={Shi, Hengcan and Dao, Son Duy and Cai, Jianfei},
  journal={International Journal of Computer Vision},
  volume={133},
  number={2},
  pages={742--759},
  year={2025},
  publisher={Springer}
}

@article{wang2025diffusion,
  title={Diffusion model is secretly a training-free open vocabulary semantic segmenter},
  author={Wang, Jinglong and Li, Xiawei and Zhang, Jing and Xu, Qingyuan and Zhou, Qin and Yu, Qian and Sheng, Lu and Xu, Dong},
  journal={IEEE Transactions on Image Processing},
  year={2025},
  publisher={IEEE}
}

@article{xie2023sed,
  title={SED: A Simple Encoder-Decoder for Open-Vocabulary Semantic Segmentation},
  author={Xie, Bin and Cao, Jiale and Xie, Jin and Khan, Fahad Shahbaz and Pang, Yanwei},
  journal={arXiv preprint arXiv:2311.15537},
  year={2023}
}

@article{cho2023cat,
  title={Cat-seg: Cost aggregation for open-vocabulary semantic segmentation},
  author={Cho, Seokju and Shin, Heeseong and Hong, Sunghwan and An, Seungjun and Lee, Seungjun and Arnab, Anurag and Seo, Paul Hongsuck and Kim, Seungryong},
  journal={arXiv preprint arXiv:2303.11797},
  year={2023}
}

@inproceedings{xu2023side,
  title={Side adapter network for open-vocabulary semantic segmentation},
  author={Xu, Mengde and Zhang, Zheng and Wei, Fangyun and Hu, Han and Bai, Xiang},
  booktitle={Proceedings of the IEEE/CVF Conference on Computer Vision and Pattern Recognition},
  pages={2945--2954},
  year={2023}
}

@inproceedings{xu2022simple,
  title={A simple baseline for open-vocabulary semantic segmentation with pre-trained vision-language model},
  author={Xu, Mengde and Zhang, Zheng and Wei, Fangyun and Lin, Yutong and Cao, Yue and Hu, Han and Bai, Xiang},
  booktitle={European Conference on Computer Vision},
  pages={736--753},
  year={2022},
  organization={Springer}
}

@inproceedings{khattak2023maple,
  title={Maple: Multi-modal prompt learning},
  author={Khattak, Muhammad Uzair and Rasheed, Hanoona and Maaz, Muhammad and Khan, Salman and Khan, Fahad Shahbaz},
  booktitle={Proceedings of the IEEE/CVF Conference on Computer Vision and Pattern Recognition},
  pages={19113--19122},
  year={2023}
}

@article{zang2022unified,
  title={Unified vision and language prompt learning},
  author={Zang, Yuhang and Li, Wei and Zhou, Kaiyang and Huang, Chen and Loy, Chen Change},
  journal={arXiv preprint arXiv:2210.07225},
  year={2022}
}

@article{bahng2022exploring,
  title={Exploring visual prompts for adapting large-scale models},
  author={Bahng, Hyojin and Jahanian, Ali and Sankaranarayanan, Swami and Isola, Phillip},
  journal={arXiv preprint arXiv:2203.17274},
  year={2022}
}

@inproceedings{zhou2022conditional,
  title={Conditional prompt learning for vision-language models},
  author={Zhou, Kaiyang and Yang, Jingkang and Loy, Chen Change and Liu, Ziwei},
  booktitle={Proceedings of the IEEE/CVF Conference on Computer Vision and Pattern Recognition},
  pages={16816--16825},
  year={2022}
}

@article{zhou2022learning,
  title={Learning to prompt for vision-language models},
  author={Zhou, Kaiyang and Yang, Jingkang and Loy, Chen Change and Liu, Ziwei},
  journal={International Journal of Computer Vision},
  volume={130},
  number={9},
  pages={2337--2348},
  year={2022},
  publisher={Springer}
}

@article{wu2021data,
  title={Data efficient language-supervised zero-shot recognition with optimal transport distillation},
  author={Wu, Bichen and Cheng, Ruizhe and Zhang, Peizhao and Vajda, Peter and Gonzalez, Joseph E},
  journal={arXiv preprint arXiv:2112.09445},
  year={2021}
}

@inproceedings{cui2022contrastive,
  title={Contrastive vision-language pre-training with limited resources},
  author={Cui, Quan and Zhou, Boyan and Guo, Yu and Yin, Weidong and Wu, Hao and Yoshie, Osamu and Chen, Yubo},
  booktitle={European Conference on Computer Vision},
  pages={236--253},
  year={2022},
  organization={Springer}
}

@inproceedings{yang2022unified,
  title={Unified contrastive learning in image-text-label space},
  author={Yang, Jianwei and Li, Chunyuan and Zhang, Pengchuan and Xiao, Bin and Liu, Ce and Yuan, Lu and Gao, Jianfeng},
  booktitle={Proceedings of the IEEE/CVF Conference on Computer Vision and Pattern Recognition},
  pages={19163--19173},
  year={2022}
}

@article{li2021supervision,
  title={Supervision exists everywhere: A data efficient contrastive language-image pre-training paradigm},
  author={Li, Yangguang and Liang, Feng and Zhao, Lichen and Cui, Yufeng and Ouyang, Wanli and Shao, Jing and Yu, Fengwei and Yan, Junjie},
  journal={arXiv preprint arXiv:2110.05208},
  year={2021}
}

@inproceedings{jia2022visual,
  title={Visual prompt tuning},
  author={Jia, Menglin and Tang, Luming and Chen, Bor-Chun and Cardie, Claire and Belongie, Serge and Hariharan, Bharath and Lim, Ser-Nam},
  booktitle={European Conference on Computer Vision},
  pages={709--727},
  year={2022},
  organization={Springer}
}

@inproceedings{rao2022denseclip,
  title={Denseclip: Language-guided dense prediction with context-aware prompting},
  author={Rao, Yongming and Zhao, Wenliang and Chen, Guangyi and Tang, Yansong and Zhu, Zheng and Huang, Guan and Zhou, Jie and Lu, Jiwen},
  booktitle={Proceedings of the IEEE/CVF Conference on Computer Vision and Pattern Recognition},
  pages={18082--18091},
  year={2022}
}

@inproceedings{luddecke2022image,
  title={Image segmentation using text and image prompts},
  author={L{\"u}ddecke, Timo and Ecker, Alexander},
  booktitle={Proceedings of the IEEE/CVF Conference on Computer Vision and Pattern Recognition},
  pages={7086--7096},
  year={2022}
}

@inproceedings{kwon2023probabilistic,
  title={Probabilistic Prompt Learning for Dense Prediction},
  author={Kwon, Hyeongjun and Song, Taeyong and Jeong, Somi and Kim, Jin and Jang, Jinhyun and Sohn, Kwanghoon},
  booktitle={Proceedings of the IEEE/CVF Conference on Computer Vision and Pattern Recognition},
  pages={6768--6777},
  year={2023}
}

@inproceedings{zhu2023segprompt,
  title={SegPrompt: Boosting Open-world Segmentation via Category-level Prompt Learning},
  author={Zhu, Muzhi and Li, Hengtao and Chen, Hao and Fan, Chengxiang and Mao, Weian and Jing, Chenchen and Liu, Yifan and Shen, Chunhua},
  booktitle={Proceedings of the IEEE/CVF International Conference on Computer Vision},
  pages={999--1008},
  year={2023}
}

@inproceedings{liang2023open,
  title={Open-vocabulary semantic segmentation with mask-adapted clip},
  author={Liang, Feng and Wu, Bichen and Dai, Xiaoliang and Li, Kunpeng and Zhao, Yinan and Zhang, Hang and Zhang, Peizhao and Vajda, Peter and Marculescu, Diana},
  booktitle={Proceedings of the IEEE/CVF Conference on Computer Vision and Pattern Recognition},
  pages={7061--7070},
  year={2023}
}

@inproceedings{yu2023zero,
  title={Zero-shot Referring Image Segmentation with Global-Local Context Features},
  author={Yu, Seonghoon and Seo, Paul Hongsuck and Son, Jeany},
  booktitle={Proceedings of the IEEE/CVF Conference on Computer Vision and Pattern Recognition},
  pages={19456--19465},
  year={2023}
}

@inproceedings{pastore2021closer,
  title={A closer look at self-training for zero-label semantic segmentation},
  author={Pastore, Giuseppe and Cermelli, Fabio and Xian, Yongqin and Mancini, Massimiliano and Akata, Zeynep and Caputo, Barbara},
  booktitle={Proceedings of the IEEE/CVF Conference on Computer Vision and Pattern Recognition},
  pages={2693--2702},
  year={2021}
}

@inproceedings{cheng2021sign,
  title={Sign: Spatial-information incorporated generative network for generalized zero-shot semantic segmentation},
  author={Cheng, Jiaxin and Nandi, Soumyaroop and Natarajan, Prem and Abd-Almageed, Wael},
  booktitle={Proceedings of the IEEE/CVF International Conference on Computer Vision},
  pages={9556--9566},
  year={2021}
}

@inproceedings{gu2020context,
  title={Context-aware feature generation for zero-shot semantic segmentation},
  author={Gu, Zhangxuan and Zhou, Siyuan and Niu, Li and Zhao, Zihan and Zhang, Liqing},
  booktitle={Proceedings of the 28th ACM International Conference on Multimedia},
  pages={1921--1929},
  year={2020}
}

@article{bucher2019zero,
  title={Zero-shot semantic segmentation},
  author={Bucher, Maxime and Vu, Tuan-Hung and Cord, Matthieu and P{\'e}rez, Patrick},
  journal={Advances in Neural Information Processing Systems},
  volume={32},
  year={2019}
}

@inproceedings{xian2019semantic,
  title={Semantic projection network for zero-and few-label semantic segmentation},
  author={Xian, Yongqin and Choudhury, Subhabrata and He, Yang and Schiele, Bernt and Akata, Zeynep},
  booktitle={Proceedings of the IEEE/CVF Conference on Computer Vision and Pattern Recognition},
  pages={8256--8265},
  year={2019}
}

@article{wu2023diffumask,
  title={Diffumask: Synthesizing images with pixel-level annotations for semantic segmentation using diffusion models},
  author={Wu, Weijia and Zhao, Yuzhong and Shou, Mike Zheng and Zhou, Hong and Shen, Chunhua},
  journal={arXiv preprint arXiv:2303.11681},
  year={2023}
}

@inproceedings{ding2022decoupling,
  title={Decoupling zero-shot semantic segmentation},
  author={Ding, Jian and Xue, Nan and Xia, Gui-Song and Dai, Dengxin},
  booktitle={Proceedings of the IEEE/CVF Conference on Computer Vision and Pattern Recognition},
  pages={11583--11592},
  year={2022}
}

@inproceedings{zhou2023zegclip,
  title={Zegclip: Towards adapting clip for zero-shot semantic segmentation},
  author={Zhou, Ziqin and Lei, Yinjie and Zhang, Bowen and Liu, Lingqiao and Liu, Yifan},
  booktitle={Proceedings of the IEEE/CVF Conference on Computer Vision and Pattern Recognition},
  pages={11175--11185},
  year={2023}
}

@article{darcet2023vision,
  title={Vision transformers need registers},
  author={Darcet, Timoth{\'e}e and Oquab, Maxime and Mairal, Julien and Bojanowski, Piotr},
  journal={arXiv preprint arXiv:2309.16588},
  year={2023}
}

@inproceedings{wang2025declip,
  title={Declip: Decoupled learning for open-vocabulary dense perception},
  author={Wang, Junjie and Chen, Bin and Li, Yulin and Kang, Bin and Chen, Yichi and Tian, Zhuotao},
  booktitle={Proceedings of the Computer Vision and Pattern Recognition Conference},
  pages={14824--14834},
  year={2025}
}

@article{ppl,
  author       = {Yao Wu and
                  Mingwei Xing and
                  Yachao Zhang and
                  Yuan Xie and
                  Zongze Wu and
                  Yanyun Qu},
  title        = {Perturbed Progressive Learning for Semisupervised Defect Segmentation},
  journal      = {{IEEE} Trans. Neural Networks Learn. Syst.},
  volume       = {35},
  number       = {5},
  pages        = {6118--6132},
  year         = {2024},
}

@article{regzsl,
  author       = {Yao Wu and
                  Xia Kong and
                  Yuan Xie and
                  Yanyun Qu},
  title        = {{RE-GZSL:} Relation Extrapolation for Generalized Zero-Shot Learning},
  journal      = {{IEEE} Trans. Circuits Syst. Video Technol.},
  volume       = {35},
  number       = {3},
  pages        = {1973--1986},
  year         = {2025}
}

@inproceedings{unidseg,
  author       = {Yao Wu and
                  Mingwei Xing and
                  Yachao Zhang and
                  Xiaotong Luo and
                  Yuan Xie and
                  Yanyun Qu},
  title        = {{UniDSeg:} Unified Cross-Domain 3D Semantic Segmentation via Visual Foundation Models Prior},
  booktitle    = {Advances in Neural Information Processing Systems},
  pages      = {101223--101249},
  year         = {2024}
}

@article{ftd,
  author       = {Yao Wu and
                  Mingwei Xing and
                  Yachao Zhang and
                  Yuan Xie and
                  Yanyun Qu},
  title        = {Fusion-Then-Distillation: Toward Cross-Modal Positive Distillation for Domain Adaptive 3D Semantic Segmentation},
  journal      = {TCSVT},
  volume       = {35},
  number       = {9},
  pages        = {9030--9045},
  year         = {2025}
}

\end{sloppypar}
\end{document}